\documentclass{article} 
\usepackage{collas2022_conference,times}

\usepackage{amsmath,amsfonts,bm}

\def\eqref#1{equation~\ref{#1}}

\def\1{\bm{1}}

\DeclareMathAlphabet{\mathsfit}{\encodingdefault}{\sfdefault}{m}{sl}
\SetMathAlphabet{\mathsfit}{bold}{\encodingdefault}{\sfdefault}{bx}{n}

\usepackage{hyperref}
\hypersetup{
    colorlinks=true,
    linkcolor=red,
    filecolor=magenta,      
    urlcolor=blue,
    citecolor=purple,
    pdftitle={Overleaf Example},
    pdfpagemode=FullScreen,
    }
    
\usepackage{enumitem}
\usepackage{graphicx}
\usepackage{tikz}
\usepackage{comment}
\usepackage{amsmath,amssymb} 
\usepackage{color}
\usepackage{verbatim}
\usepackage{subcaption}
\usepackage{booktabs}
\usepackage{enumitem}
\usepackage{algorithm2e}
\usepackage{lipsum}
\usepackage{xfrac}
\usepackage{soul}
\usepackage{glossaries}
\usepackage{wrapfig,adjustbox}

\title{Online Continual Learning for Embedded Devices}

\author{Tyler~L.~Hayes$^1$, Christopher Kanan$^{1,2}$\\
Rochester Institute of Technology$^1$, University of Rochester$^2$\\
{tlh6792@rit.edu}, {ckanan@cs.rochester.edu} \\
}

\collasfinalcopy 

\begin{document}

\maketitle

\begin{abstract}
    Real-time on-device continual learning is needed for new applications such as home robots, user personalization on smartphones, and augmented/virtual reality headsets. However, this setting poses unique challenges: embedded devices have limited memory and compute capacity and conventional machine learning models suffer from catastrophic forgetting when updated on non-stationary data streams. While several online continual learning models have been developed, their effectiveness for embedded applications has not been rigorously studied. In this paper, we first identify criteria that online continual learners must meet to effectively perform real-time, on-device learning. We then study the efficacy of several online continual learning methods when used with mobile neural networks. We measure their performance, memory usage, compute requirements, and ability to generalize to out-of-domain inputs\footnote{\url{https://github.com/tyler-hayes/Embedded-CL}}.
\end{abstract}

\section{Introduction}

Continual machine learning systems have the ability to learn from ever-growing data streams~\citep{parisi2019continual}. In contrast, conventional machine learning algorithms typically assume that there is a static training and evaluation dataset. Continual learning has emerged as a popular research area. One of the most critical applications for continual learning is using it on embedded devices such as mobile phones, virtual/augmented reality (VR/AR) headsets, robots, vehicles, and smart appliances. VR headsets use continual learning to localize the position of the wearer within the boundary that the user has established so that the user does not collide with obstacles~\citep{o2020reality}. AR headsets require continual learning to identify relevant objects and regions in the field of view to appropriately position virtual perceptual information. Household robotic devices need to learn the identity of the individuals, pets, and objects in the house. Typically, inference for these applications must be done within embedded devices to minimize latency, but continual on-device learning is critical to preserving privacy and security of the user.

\begin{figure}[t]
\begin{center}
    \includegraphics[width=0.6\textwidth]{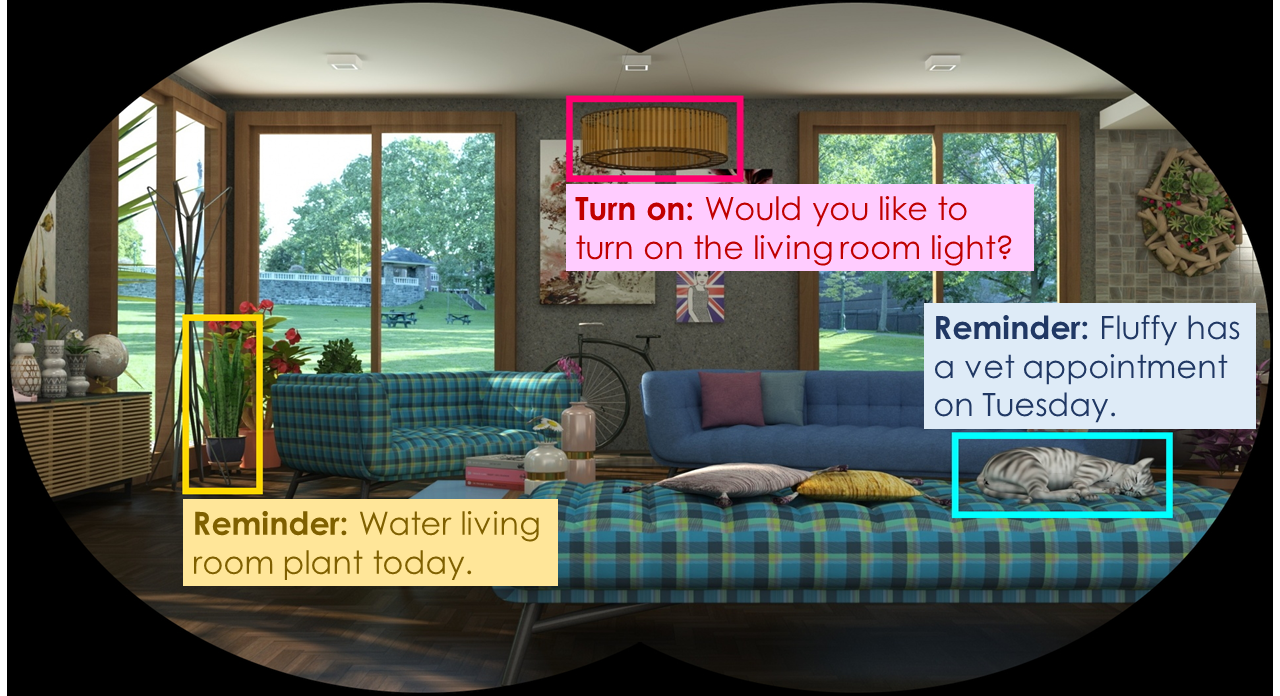}
\end{center}
\caption{Using online continual learning, an AR headset could update the locations of specific object instances in the frame of a wearer to provide real-time feedback. This requires immediate learning and inference on the compute and memory constrained headset. Moreover, no assumptions can be made about the order in which the frames of data will be provided and task labels cannot be assumed to be provided. This paper establishes the necessary criteria to support the types of AR applications demonstrated here. Moreover, we also provide initial baselines that meet these criteria.}
\label{fig:criteria}
\end{figure}

Conventional machine learning systems trained with empirical risk minimization assume that the data is independent and identically distributed (iid), which is typically enforced by shuffling the data. In continual learning, this assumption is violated, which results in catastrophic forgetting~\citep{french1999catastrophic,parisi2019continual}. Hence, the continual learning research community has focused on solving this catastrophic forgetting problem in a variety of scenarios. However, most of these scenarios do not match the conditions an agent would face for embedded applications. For embedded \emph{supervised} continual learning systems, we argue these capabilities are essential:
\begin{enumerate}[noitemsep, nolistsep]
    \item Online learning and inference in a compute and memory constrained environment.
    \item The ability to learn from data in any order without catastrophic forgetting, including learning from iid (shuffled) data streams, streams ordered by category, and everything in between.
    \item Making no assumptions about the availability of task labels during inference.
    \item Efficiently learning and generalizing with as few labeled examples as possible.
\end{enumerate}
These criteria can be summarized as stating that we need sample efficient, order agnostic, online continual learning (see Fig.~\ref{fig:criteria} for a motivating example).

\textbf{In this paper, we study continual learning for embedded devices. We make the following contributions:}
\begin{enumerate}[noitemsep, nolistsep]
    \item We establish the criteria needed for continual learning on embedded devices, a real-world problem where continual learning is needed.
    
    \item We compare seven algorithms for online learning when used with five convolutional neural networks (CNNs) designed for embedded devices based on their ability to learn from both shuffled data and data sorted by category, which are the extreme best and worst case scenarios, respectively.
    
    \item We conduct experiments on three high-resolution image classification datasets: OpenLORIS, Places-365, and Places-Long-Tail. Algorithms are compared based on their ability to learn over time as well as their computational and memory requirements.
    
    \item We make baseline recommendations for future researchers based on each online continual learner's ability to learn from temporally correlated video streams, learn/generalize from very few labeled examples, scale to large-scale data streams with hundreds of classes and millions of examples, and perform well on imbalanced data streams.
    
\end{enumerate}

\section{Problem Formulation}

We study \emph{online continual learning} from possibly infinite data streams. That is, a model learns a dataset $\mathcal{D}=\left\{\left(\mathbf{X}_{t}, y_{t}\right)\right\}_{t=1}^{\infty}$ one instance at a time, where $\left(\mathbf{X}_{t}, y_{t}\right)$ is an example/label pair received at time $t$. In our paper, all examples are images. Each example is seen only once, where it cannot be revisited unless it is cached in auxiliary memory. The online continual learning setting is needed because many embedded applications receive a limited amount of data and only have a limited amount of time to process the data. For example, a user might use their smartphone to take a handful of photos of objects in their home. A classification model would then be required to rapidly process these few object instances to provide a prediction in real-time. Moreover, many embedded systems collect data in a natural streaming setting (e.g., video cameras collect data frame-by-frame) and it would be ideal to process this data in the real-time order in which it was received.

For online continual learners to be deployed on embedded devices, they must also operate under memory and time constraints to operate on-device. The learner must be performant regardless of the ordering of the instances. This is because an agent operating in the natural world might not have direct control over the order in which visual inputs are received and cannot assume they will be randomly shuffled (iid). Further, agents can only assume access to an input $\mathbf{X}_{t}$ for inference (i.e., labels indicating which task an input came from may not be available). This is because providing task labels to agents requires additional annotation efforts, which may not be available for some applications. Finally, to learn quickly, agents must be able to generalize from very few examples.

\section{Related Work}


Continual learning involves training agents over time from non-stationary data streams. However, one challenge with updating agents on non-stationary data distributions is catastrophic forgetting of previous knowledge~\citep{mccloskey1989,french1999catastrophic}. To overcome forgetting, researchers have studied regularization techniques~\citep{aljundi2018memory,chaudhry2018riemannian,chaudhry2019efficient,dhar2019learning,kirkpatrick2017,li2016learning,lopez2017gradient,ritter2018online,serra2018overcoming,zenke2017continual}, network expansion methods~\citep{hayes2019lifelong,hou2018lifelong,ostapenko2019learning,rusu2016progressive,yoon2018lifelong}, and experience replay~\citep{hayes2021replay,belouadah2019il2m,castro2018end,chaudhry2019efficient,douillard2020podnet,french1997pseudo,hayes2019memory,hayes2019remind,hou2019unified,kemker2018fearnet,rebuffi2016icarl,tao2020topology,wu2019large}. Regularization techniques add a constraint to the loss function to constrain parameter updates. Network expansion methods add new neurons to the network as new tasks become available. Experience replay methods maintain a subset of previous data in a memory buffer and mix old data with new data to prevent forgetting.

\paragraph{Continual Embedded Learning.}
While many continual learning methods have been developed to overcome forgetting, some make strong assumptions which make them ill-suited for embedded applications. For example, many algorithms train using large batches consisting of thousands of training examples and if trained instance-by-instance, their performance degrades~\citep{hayes2019remind}. Batch learning is undesirable for embedded applications since it requires time to queue up batches and more time to train. For example, batch learning methods in \citet{hayes2019lifelong} required over 60 hours to train, while the proposed online learning method required only 30 minutes. Further, while looping over batches to train, the model cannot make inferences, which can be prohibitive in real-time scenarios. Moreover, some algorithms require task labels to be provided during inference and if this assumption is violated, their performance also degrades~\citep{hayes2019lifelong,hayes2019remind}. Also relevant to this work are methods designed specifically for few-shot continual learning~\citep{ayub2020cognitively,ayub2021fsiol,tao2020few,tao2020topology}. These methods consider the ability of algorithms to generalize from very few instances, which is one of the necessary capabilities for continual learning on embedded devices. In this work, we focus on studying online continual learning algorithms that learn instance-by-instance in a single pass through the dataset, with memory and compute constraints, and without the need for task labels during inference. We believe this setup is representative of how agents would be trained and evaluated on embedded devices. 

There has been a limited amount of research studying continual learning for embedded applications~\citep{pellegrini2020latent,pellegrini2021continual,demosthenous2021continual,li2019rilod}. Most focus on supervised classification using small batches on the CORe50 dataset~\citep{lomonaco2017core50} with a MobileNet-v1 backbone~\citep{pellegrini2020latent,pellegrini2021continual,demosthenous2021continual}, while one focuses on continual object detection~\citep{li2019rilod}. Small batches do not meet real-world on-device continual learning needs, where a user wants to minimize the number of samples. To address this, we study seven online continual learning algorithms that learn instance-by-instance. Each method is combined with five different backbone CNNs. We evaluate the performance of these methods on three unique datasets that test the robustness of algorithms to scale (i.e., resolution, number of classes, and number of training examples), imbalanced/long-tailed data streams, and temporally correlated video streams. Existing work has separately evaluated the robustness of continual learners to scale~\citep{hou2019unified,rebuffi2016icarl,castro2018end,wu2019large}, video streams~\citep{hayes2019remind,hayes2019lifelong,pellegrini2020latent,pellegrini2021continual}, imbalanced data streams~\citep{aggarwal2021comparative,belouadah2020active,hu2020learning}, and low-shot learning~\citep{ayub2020cognitively, ayub2021fsiol,tao2020few,tao2020topology}. Conversely, we provide the first comprehensive study that directly compares the robustness of online learners to each of these scenarios when paired with CNNs specifically designed for on-device learning. We compare methods in terms of classification efficacy, memory, and compute.

\paragraph{Efficient On-Device Learning.}

Several convolutional neural network (CNN) architectures have been designed to address the need for on-device learning. These architectures are designed to achieve high accuracy while also considering computational efficiency. Methods that have been proposed for improving efficiency and reducing memory (e.g., network parameters) include 1$\times$1 convolutions (SqueezeNet)~\citep{iandola2016squeezenet}, group convolutions (ShuffleNet, CondenseNet)~\citep{zhang2018shufflenet,huang2018condensenet}, depth-wise separable convolutions (MobileNet-v1, MobileNet-v2)~\citep{howard2017mobilenets,sandler2018mobilenetv2}, and point-wise convolutions (ShiftNet)~\citep{wu2018shift}. Memory can also be reduced using pruning~\citep{alvarez2016learning,han2015learning,hu2016network,lecun1989optimal,li2016pruning,louizos2017learning,srinivas2015data}, quantization~\citep{courbariaux2016binarized,jacob2018quantization,kim2016bitwise}, model compression~\citep{jacob2018quantization,krishnamoorthi2018quantizing,wu2016quantized,soudry2014expectation,zhou2017incremental,zhou2016dorefa,rastegari2016xnor}, or network distillation~\citep{bucilua2006model,hinton2015distilling}. Beyond this, reinforcement learning and neural architecture search have been used to find architectures that balance accuracy with efficiency~\citep{zoph2016neural,zoph2018learning,baker2016designing,liu2018progressive,pham2018efficient,tan2019mnasnet,cai2018proxylessnas,yang2018netadapt}. In this paper, we use two common architectures designed to fit on mobile devices that achieve strong performance on supervised ImageNet classification: MobileNet-v3~\citep{howard2019searching} and EfficientNet~\citep{tan2019efficientnet}, which we discuss more in Sec.~\ref{sec:architectures}.

\section{Experimental Setup}
\label{sec:experimental-setup}

We incrementally train a neural network $\hat{y}_t=F\left(G\left(\mathbf{X}_{t}\right)\right)$ using supervised online continual learning, where $\mathbf{X}_{t}$ is an image at time $t$ and $\hat{y}_{t}$ is the predicted label. Following traditional transfer learning setups, $G\left(\cdot\right)$ is a backbone CNN pre-trained using supervised learning on the ImageNet-1k dataset~\citep{russakovsky2015imagenet} and $F\left(\cdot\right)$ is a classifier. We assume the output of $G\left(\cdot\right)$ is a vector, which can be obtained using global average pooling or other pooling mechanisms. In our setup, we freeze $G\left(\cdot\right)$ after pre-training and train $F\left(\cdot\right)$ using online updates. That is, we focus on updating the classifier in an online fashion (i.e., one example at a time) using fixed feature representations. While updating features could improve performance further, this requires additional compute time, which is undesirable in embedded settings where training and inference must happen with limited compute capacity. Using this setup, we study five backbone architectures for $G\left(\cdot\right)$ and seven online continual learning methods to update $F\left(\cdot\right)$, which we describe next.

\subsection{Network Architectures}
\label{sec:architectures}

We study five backbone CNN architectures, which were chosen due to their small size, efficiency, and competitive performance when trained using supervised learning on ImageNet-1k.

\textbf{MobileNet-v3}~\citep{howard2019searching} builds on two previous MobileNet architectures designed for mobile and embedded applications~\citep{howard2017mobilenets,sandler2018mobilenetv2}. Specifically, it combines depth-wise separable convolutions, inverted residual connections, squeeze and excitation attention modules~\citep{iandola2016squeezenet}, neural architecture search (NAS), and more efficient nonlinearities. There are two versions of the CNN: \emph{MobileNet-v3 (Small)}, which uses fewer resources and \emph{MobileNet-v3 (Large)}, which uses more resources, but achieves higher classification accuracies. We study both versions of the architecture. The small architecture contains 927,000 parameters and achieves a top-1 accuracy of 67.7\% on ImageNet-1k. The large architecture contains 2.91 million parameters and achieves a top-1 accuracy of 74.0\% on ImageNet-1k.
    
\textbf{EfficientNet}~\citep{tan2019efficientnet} uses fixed scaling coefficients to scale the width, depth, and resolution of a network with the amount of computational resources available. Similar to MobileNets, EfficientNets combine inverted residual connections and squeeze and excitation modules with NAS to find an architecture that balances accuracy with efficiency. We use the two smallest EfficientNets: \emph{EfficientNet-B0} (4 million parameters) and \emph{EfficientNet-B1} (6.5 million parameters). EfficientNet-B0 achieves a top-1 accuracy of 77.7\% on ImageNet-1k, while EfficientNet-B1 achieves a top-1 accuracy of 78.6\%.
    
\textbf{ResNet-18}~\citep{He_2016_CVPR} is the smallest ResNet architecture commonly used in practice. This architecture was chosen since it is a common architecture used in continual learning literature~\citep{rebuffi2016icarl,castro2018end,hayes2019remind,douillard2020podnet,wu2019large}. It contains 11 million parameters and achieves a top-1 accuracy of 69.8\% on ImageNet-1k.

For all architectures, we global average pool features prior to the classifier. This yields features of the following dimensions: MobileNet-v3 (Small) 576-d, MobileNet-v3 (Large) 960-d, EfficientNet-B0 1280-d, EfficientNet-B1 1280-d, and ResNet-18 512-d.

\subsection{Online Continual Learning Models}
\label{sec:online-models}

We study seven online continual learning methods for updating the classifier $F\left(\cdot\right)$ using pre-trained universal image features from $G\left(\cdot\right)$. These methods were chosen due to their ability to learn one sample at a time in a single pass over a dataset without task labels (i.e., online continual learning) with low memory and compute requirements.

\textbf{Fine-Tune} incrementally fine-tunes a fully-connected output layer one sample at a time using stochastic gradient descent and a cross-entropy loss. This method does not have any mechanisms to prevent catastrophic forgetting.

\textbf{Nearest Class Mean} (NCM) maintains one running mean vector per class (i.e., $\mathbf{w}_{k}$ is the mean for the $k$-th class), each with an associated counter denoting the number of samples represented in each mean ($c_{k}$). Given a new data vector $\mathbf{x}_{t}$ with associated label $y_{t}$ at time $t$, the class mean and associated counter are updated as:
\begin{equation}
\label{eq:running-mean}
\begin{split}
    \mathbf{w}_{y_{t}} \leftarrow \frac{c_{y_t}\mathbf{w}_{y_{t}}+\mathbf{x}_{t}}{c_{y_{t}}+1} \enspace, \qquad c_{y_{t}} \leftarrow c_{y_{t}} + 1 \enspace .
\end{split}
\end{equation}
During inference, it assigns the label of the nearest class mean to a new example. This is a common baseline in continual learning~\citep{rebuffi2016icarl}. Following past work we use Euclidean distance for the metric.

\textbf{Streaming One-vs-Rest} (SOvR) compares how close a new example is to a class mean vector while also considering its distance to data from other classes, which is reminiscent of support vector machines. Specifically, it maintains one running mean vector and associated count per class, which are updated using Eq.~\ref{eq:running-mean}. During inference, the method computes one vector per class ($\mathbf{\tilde{w}}_{k}$) that is representative of the mean of data not belonging to that class, i.e.,
\begin{equation}
    \mathbf{\tilde{w}}_{k} \leftarrow \frac{1}{N}\sum_{i\neq k}{c_{i}\mathbf{w}_{i}} \enspace ,
\end{equation}
where $N=\sum_{i} c_{i}$ is the sum of all class counts. To assign a label to a new example, the SOvR method first computes the distance of the new sample to each vector $\mathbf{w}_{k}$ as $d_{k}$ and to each vector $\mathbf{\tilde{w}}_{k}$ as $\tilde{d}_{k}$ by computing the dot product between the new sample with each of the vectors. The score for class $k$ is then computed as $s_{k}=\frac{d_{k}}{d_{k} + \tilde{d}_{k}}$ and a label is assigned by taking the argmax over all $s_{k}$. 

\textbf{Streaming Linear Discriminant Analysis} (SLDA) maintains one running mean vector per class with an associated counter, which are updated using Eq.~\ref{eq:running-mean}, and one shared running covariance matrix among classes. We use the implementation from \citet{hayes2019lifelong} to update the covariance matrix and compute predictions. This implementation was shown to work well on the large-scale ImageNet-1k dataset. Intuitively, SLDA makes predictions by assigning a new example the label of the closest Gaussian in feature space defined using the running class means and shared covariance matrix. NCM is a special case of SLDA where the covariance matrix is equal to the identity matrix.

\textbf{Streaming Gaussian Naive Bayes} is related to Streaming Quadratic Discriminant Analysis (SQDA). While SQDA stores one running covariance matrix per class, Gaussian Naive Bayes instead stores one variance vector per class (i.e., diagonal covariance matrices that assume independent features). It also stores one running mean vector per class with an associated counter. The advantage of Gaussian Naive Bayes to SQDA is that storing diagonal covariance matrices requires significantly less memory than storing full covariance matrices. Compared to SLDA, Gaussian Naive Bayes is able to more accurately represent the variance for each class instead of using a shared covariance matrix. However, unlike SLDA and SQDA, Gaussian Naive Bayes assumes feature independence with its variance vectors, which is not always guaranteed in practice. We use Eq.~\ref{eq:running-mean} to update the running mean vectors and Welford's sample variance algorithm~\citep{welford1962note} to update the running variance vectors in a numerically stable way. We perform inference using the same procedure as SLDA but with the per class variance vectors (i.e., diagonal covariance matrices) instead of a shared covariance matrix.

\textbf{Online Perceptron} maintains one weight vector for each class. The first time a sample from a new class is seen, the weight vector for the class is set to the sample. After that, each time the model misclassifies a new sample $\mathbf{x}_{t}$ with label $y_{t}$, the associated class weight vector is updated as:
\begin{equation}
    \mathbf{w}_{y_{t}} \leftarrow \mathbf{w}_{y_{t}} + \mathbf{x}_{t} \enspace ,
\end{equation}
and the weight vector of the incorrect class with the highest score ($\mathbf{w}_{i}$) is updated as
\begin{equation}
    \mathbf{w}_{i} \leftarrow \mathbf{w}_{i} - \mathbf{x}_{t} \enspace .
\end{equation}
During inference, a score for class $k$ is computed by taking the dot product between $\mathbf{w}_{k}$ and an input vector $\mathbf{x}_{t}$. A label is then assigned by taking the argmax over the computed scores for all $k$ classes. 

\textbf{Replay} mitigates catastrophic forgetting by storing a subset of previous examples in a memory replay buffer and has grown in popularity due to its strong performance on large-scale continual learning benchmarks~\citep{rebuffi2016icarl,castro2018end,hayes2019remind,douillard2020podnet}. Specifically, replay maintains a memory buffer of size $B$, which equally distributes examples among classes seen so far. When the buffer is full, it randomly replaces an example from the most represented class with a new example. During training, it randomly selects $P$ examples from the replay buffer, combines them with the new example, and makes a single update using stochastic gradient descent with a cross-entropy loss. Consistent with the other algorithms, we use replay to train a single fully-connected output layer. While effective, replay can be memory intensive due to the storage of its memory buffer.

\subsection{Datasets}
\label{sec:datasets-orders}

We use three high-resolution image classification datasets to evaluate the online continual learners.

\begin{figure}[t]
\begin{center}
    \includegraphics[width=\textwidth]{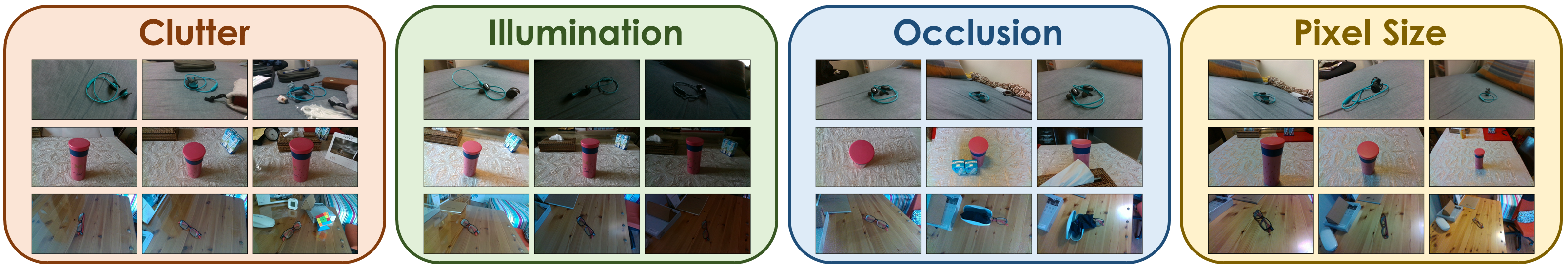}
\end{center}
\caption{Example images from the OpenLORIS video dataset~\citep{she2020openloris}. We demonstrate the varied difficulty across each of the four domain factors for three unique object instances (headphones, bottle, and glasses).}
\label{fig:open-loris}
\end{figure}

\textbf{OpenLORIS} is a dataset containing videos of different household objects taken with a camera attached to a robot~\citep{she2020openloris}. Specifically, it contains 40 unique object classes each with videos of 1 to 9 different object instances. There are a total of 121 object instances. Each object instance was recorded under four varying domain factors: amount of clutter, amount/source of illumination, amount of occlusion, and pixel size of the object in frame. For each domain, the object instances were recorded under 9 different sessions of varying difficulty, yielding 36 unique videos for each object instance. The dataset contains over 440,000 training images and over 53,000 test images. This dataset is the most realistic for our setup because it requires agents to learn from non-stationary (temporally correlated) video streams and adapt to changing domains. Example images from the OpenLORIS dataset are in Fig.~\ref{fig:open-loris}.

\textbf{Places-365} is a scene classification dataset containing over 1.8 million static images from 365 unique classes~\citep{zhou2017places}. We use the small image version of the dataset where images have been resized to 256$\times$256 pixels. We evaluate on the full validation set because the test set is behind a server. The train set contains between 3,068 and 5,000 images per class, while the validation set contains 100 images per class. We use Places-365 to evaluate how well models scale in performing continual learning on over one million high-resolution images from hundreds of classes.

\textbf{Places-Long-Tail} (Places-LT) is a long-tailed version of the static Places-365 dataset which is used to test how well algorithms work in imbalanced regimes~\citep{liu2019large}. It contains 62,500 total training images with 5 to 4,980 images per class and 365 total classes. Similar to Places-365, we use the validation set for testing. We use Places-LT to test the robustness of online continual learners to heavily imbalanced datasets, which are more representative of the distribution of objects in the natural world.

\subsection{Data Orderings}
Since the order in which data is presented to an online continual learner can potentially impact its performance, we study several different orderings of each dataset. Since OpenLORIS contains videos, we study two variants of the \emph{instance} data ordering from \citet{hayes2019memory}. The instance ordering presents videos of shuffled object instances to the learner one at a time. For example, the learner could learn a video of mug \#1, then a video of hat \#3, then a video of mug \#2, and so on. The learner is then evaluated on the full test set. The instance ordering is the most realistic ordering used in this paper since an agent could be expected to learn from temporally correlated video streams with the potential of revisiting previous classes. While the standard instance ordering presents \emph{all} videos from the dataset to the learner, we also study a smaller variant of the instance ordering that we term ``low-shot instance.'' In the \emph{low-shot instance} ordering, the learner is presented with a single video of a single object instance from each category in the dataset. After each instance is learned, the learner is evaluated on all test data belonging to the classes seen so far. Since the learner is only trained on a single video from each category and evaluated on the entire test set for the category, it must be capable of generalizing to out-of-domain inputs to perform well. This is because the test set contains videos of the same object instance under different domains (e.g., clutter, occlusion, illumination, or size of object in frame), as well as videos of other object instances from the same class under different domains. Specifically, the low-shot instance ordering of OpenLORIS requires models to learn from naturally temporally correlated data streams, generalize to different object instances from the same classes, generalize to unseen domains, and learn from very little labeled data. This setting mimics a practical setting where a user might have a particular pet cat in their home that they would like a classifier to identify. If the user only provides the classifier with a few images of the cat on a couch, it would be ideal if the classifier could still identify the same pet cat in other domains (i.e., not sitting on the couch). Moreover, in addition to using the low-shot instance ordering of OpenLORIS to understand how well each continual learner can generalize from very little labeled data, we also perform experiments on the F-SIOL-310 (Few-Shot Incremental Object Learning) dataset~\citep{ayub2021fsiol} designed specifically for few-shot continual learning in supplemental materials.

Since Places-365 and Places-LT contain static images, we study two data orderings: \emph{iid} where all of the images are randomly shuffled and \emph{class-iid} where all of the images are organized by class, but shuffled within each class. In class-iid, classes are also shuffled. The iid ordering should be easiest since machine learning models typically assume data is shuffled and models are exposed to the same classes at multiple points during training. In contrast, the non-stationarity of the class-iid ordering tests the ability of models to overcome catastrophic forgetting and is a common benchmark in continual learning~\citep{rebuffi2016icarl,castro2018end,hou2019unified,wu2019large}. Models that suffer from catastrophic forgetting typically struggle with the class-iid ordering, but perform well on the iid ordering.

\begin{table}[t]
	\begin{minipage}{0.6\linewidth}
    \begin{center}
      \includegraphics[width=\linewidth]{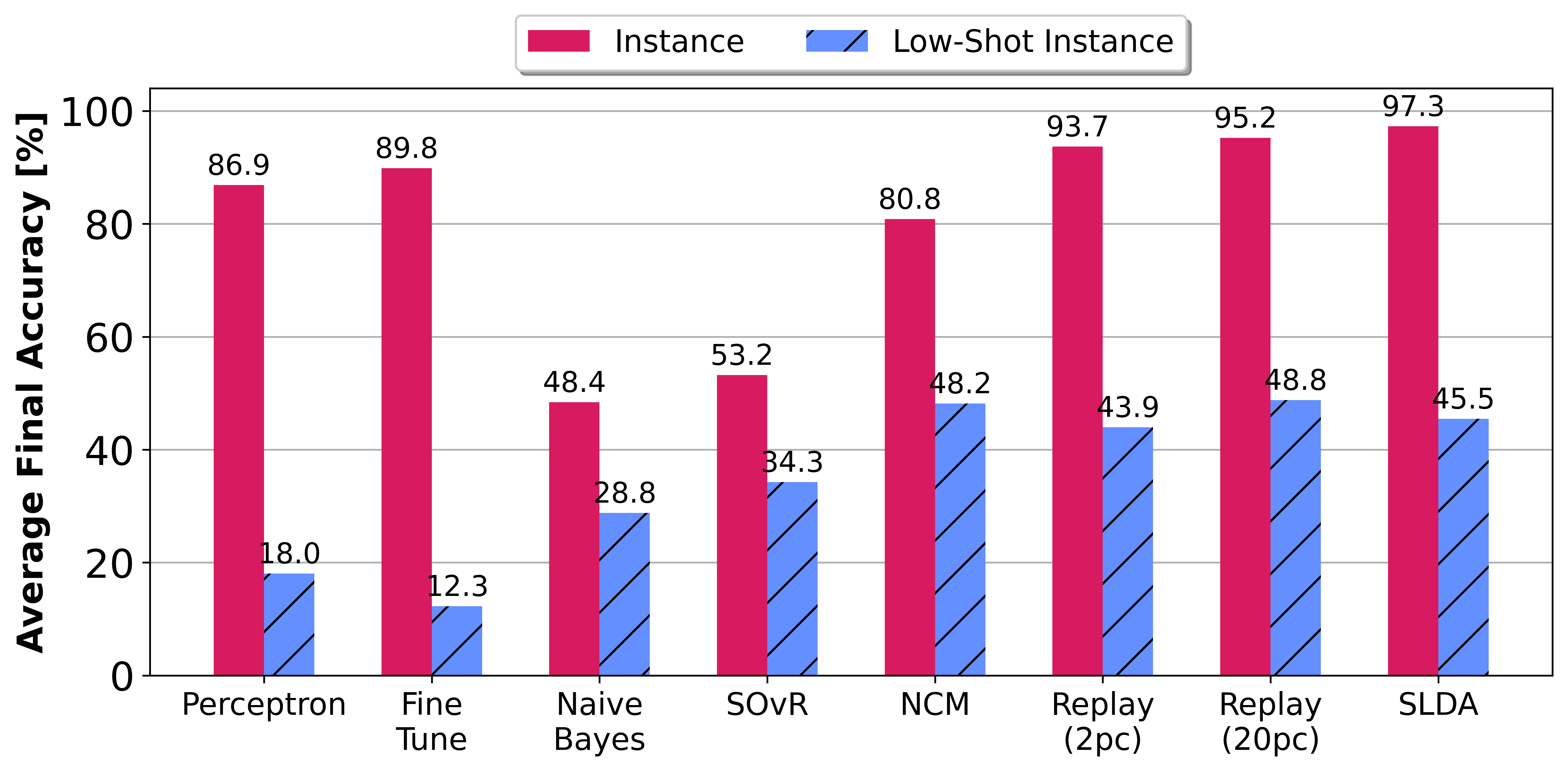}
    \end{center}
		\captionof{figure}{Final accuracy (\%) summary statistics aggregated across CNN architectures to compare online continual learning methods when trained on the OpenLORIS dataset using the instance and low-shot instance data orderings.}
		\label{fig:open-loris-bar} 
	\end{minipage}\hfill
	\begin{minipage}{0.38\linewidth}
    \begin{center}
      \includegraphics[width=\linewidth]{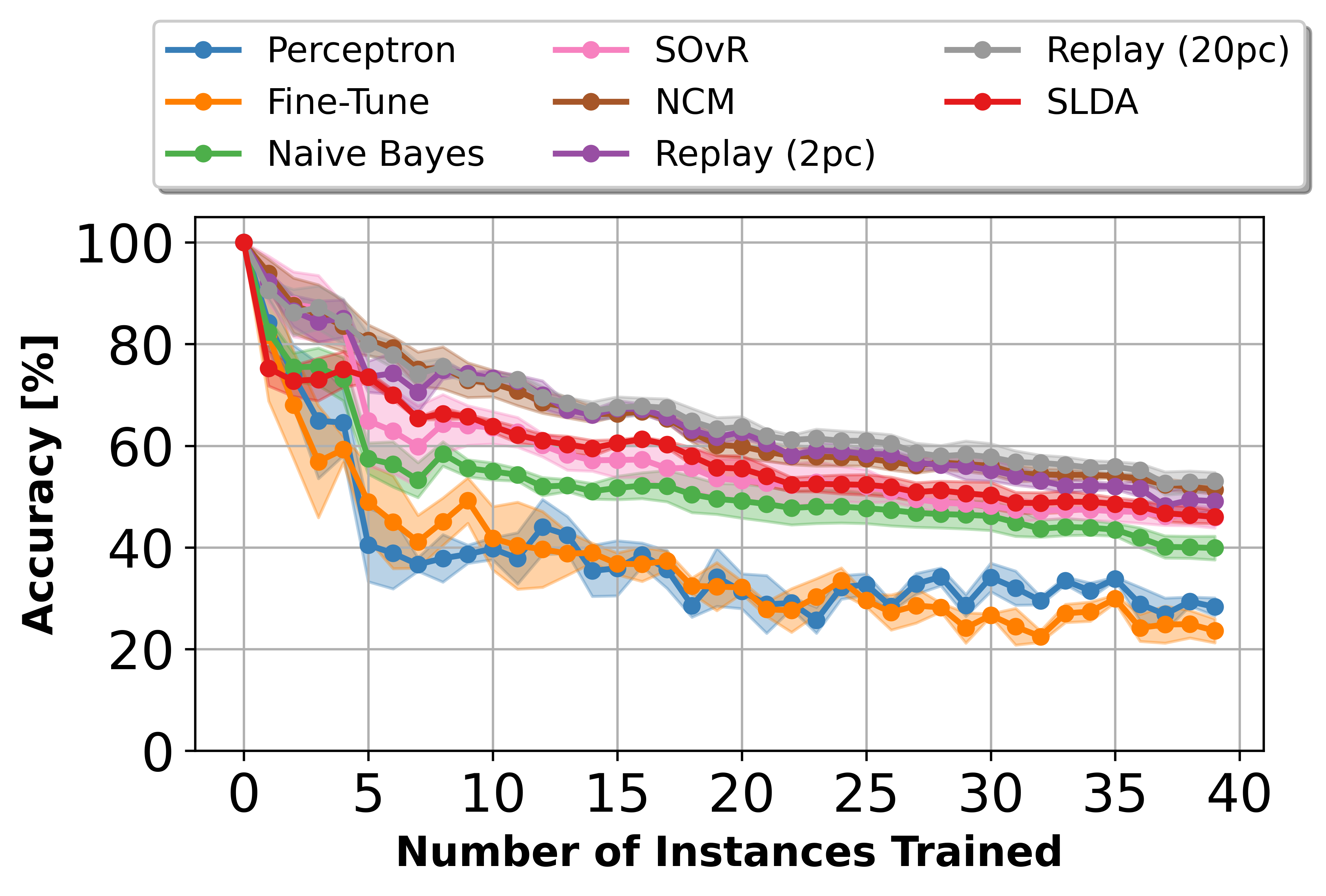}
    \end{center}
		\captionof{figure}{Learning curves of each online continual learner with the EfficientNet-B1 CNN when evaluated using the low-shot instance ordering of the OpenLORIS dataset. Each plot is the average over 3 runs with different instance permutations with standard error denoted by the shaded regions.}
		\label{fig:open-loris-learning-curve} 
	\end{minipage}
\end{table}

\subsection{Performance Metrics}
\label{sec:metrics}

We evaluate online continual learner performance on three axes: classification efficacy, memory, and compute. For classification efficacy, we use a learner's final accuracy. For scoring methods, we adopt a modified variant of the NetScore from \citet{wong2019netscore}, which provides a single score that combines accuracy, total number of parameters, and amount of compute. It is useful to compute a single metric that considers all three factors in order to better analyze the applicability of each continual learner for embedded applications. For example, consider one learner that achieves strong classification efficacy using a lot of memory and compute and consider another learner that achieves similar classification efficacy using much less memory and compute. For embedded applications, the learner that uses less memory and compute would be desirable.

The NetScore metric takes inspiration from the decibel scale used in signal processing, which compares the ratio of one value as compared to another on a logarithmic scale. Specifically, the NetScore for online learners $\mathcal{M}$ is:
\begin{equation}
    \label{eq:netscore}
    \Omega\left(\mathcal{M}\right)=s\log\left(\frac{a\left(\mathcal{M}\right)^{\alpha}}{p\left(\mathcal{M}\right)^{\beta}c\left(\mathcal{M}\right)^{\gamma}}\right) \enspace ,
\end{equation}
where $a\left(\mathcal{M}\right)$ is a method's final accuracy, $p\left(\mathcal{M}\right)$ is the total number of parameters required to store both the CNN and the online continual learner, $c\left(\mathcal{M}\right)$ is the number of seconds required to run the experiment, and $\alpha$, $\beta$, and $\gamma$ are hyper-parameters that control the influence that accuracy, memory, and compute have on the NetScore $\Omega$, respectively. Following \citet{wong2019netscore}, we set $\alpha=2$ to emphasize the importance of high classification accuracy, and $s=20$. We set $\beta=\gamma=0.25$ due to the large scale of $p\left(\mathcal{M}\right)$ and $c\left(\mathcal{M}\right)$ to ensure mostly non-negative values of $\Omega$. We include NetScore values using the original parameters from \citet{wong2019netscore} in supplemental materials.

\section{Results}
\label{sec:results}

We provide all implementation details in supplemental materials. For the OpenLORIS and Places-LT datasets, we run each online continual learner with three permutations of each data ordering and report the average performance over runs. Due to the large size of the Places-365 dataset, we run each online continual learning method with each ordering once. Note that the SOvR and NCM classifiers are agnostic to these data order permutations since their running mean estimates are unaffected by data order. We study two replay buffer sizes: one that stores 20 examples per class (20pc) and one that stores 2 examples per class (2pc) (see supplemental materials for more details).

\subsection{Results on OpenLORIS}

\begin{table}[t]
\caption{$\Omega$ NetScore values for each online continual learner using the \emph{low-shot instance} ordering of the OpenLORIS dataset. 
We also report the mean accuracy (\%) aggregated across CNN architectures.
We format the \underline{\textbf{first}}, \textbf{second} and \underline{third} best overall results.
\label{tab:open-loris-summarized-results}}
\centering
\begin{tabular}{lcccccc}
\toprule
\textsc{Method} & \textsc{MNet-S} & \textsc{MNet-L} & \textsc{ENet-B0} & \textsc{ENet-B1} & \textsc{RN-18} & \textsc{Mean} \\ 
\midrule
Perceptron & -12.3 & 3.0 & 20.6 & 18.6 & -31.8 & -0.4 \\
Fine-Tune & -45.5 & -33.4 & 15.2 & 11.4 & -71.8 & -24.8 \\
Naive Bayes & 21.1 & 33.4 & 36.9 & 31.6 & -87.4 & 7.1 \\
SOvR & 25.1 & 27.6 & 39.1 & 37.2 & 6.5 & 27.1 \\
NCM & 48.0 & 44.9 & 46.3 & 42.5 & 37.4 & \underline{\textbf{43.8}} \\
Replay (2pc) & 44.2 & 40.9 & 44.4 & 40.6 & 28.7 & 39.8 \\
Replay (20pc) & 46.7 & 44.2 & 46.1 & 43.0 & 35.8 & \textbf{43.2} \\
SLDA & 46.8 & 41.8 & 40.9 & 37.0 & 35.4 & \underline{40.4} \\
\bottomrule
\end{tabular}
\end{table}

We first studied the performance of online continual learners on the realistic OpenLORIS video dataset under two data orderings: instance and low-shot instance. Recall in the instance ordering that each learner is trained on \emph{all} object instance videos, while the low-shot instance ordering only provides the learners with a single object instance video from each of the 40 classes. To directly compare how these orderings affect model performance, we aggregated the final top-1 accuracy scores of each online continual learner across all CNN architectures in Fig.~\ref{fig:open-loris-bar}. As expected, all models perform worse when trained using the low-shot instance ordering. However, the performance differences for some models are much larger than others. For example, the perceptron and fine-tune models achieve strong performance when trained on all object instance videos, but perform poorly when only trained on a single object instance from each class. This indicates that these models exhibit poor generalization to out-of-domain inputs. While the Naive Bayes and SOvR methods outperform the perceptron and fine-tune methods in the low-shot instance setting, their performance in the full instance setting is the poorest among all methods. Overall, the replay (20pc) and SLDA methods strike the best compromise between strong performance when trained using all object instances and ability to generalize well when trained with very few instances. While replay (2pc) performs worse than replay (20pc), it still achieves competitive performance with replay (20pc) and SLDA, while requiring less memory.

To study performance differences among learners in the low-shot instance setting further, we show learning curves for each method with the EfficientNet-B1 backbone in Fig.~\ref{fig:open-loris-learning-curve}. Learning curves with additional CNN backbones are in supplemental materials. Specifically, we show the top-1 accuracy of each method on classes seen so far after viewing each object instance video. Overall, the replay (20pc), replay (2pc), and NCM methods perform consistently the best across all training increments. While results vary slightly across CNN architectures, the replay (20pc) and NCM methods are consistently the top performers, indicating that these two methods generalize the best to out-of-domain inputs (i.e., object instances filmed under alternative clutter, occlusion, illumination, or pixel size settings).

\subsubsection{NetScore Performance}

Online continual learning on embedded devices poses unique challenges to learners; they must strike a balance between achieving strong performance, while also operating under memory and compute constraints. To better understand the practical usage of each method, we report the $\Omega$ NetScore values of continual learners with each CNN in Table~\ref{tab:open-loris-summarized-results}. All methods were evaluated on the same hardware for consistency. Overall, the NCM method performed best, followed by replay (20pc) and SLDA. Since the NCM method only requires the storage and updates of class mean vectors, it is both computationally and memory efficient. While replay (20pc) and SLDA achieve strong performance in Fig.~\ref{fig:open-loris-bar}, their storage requirements and compute times are longer than NCM, meaning they have lower NetScores. Fine-tune performs the worst overall, followed by the perceptron and Naive Bayes. Interestingly, Naive Bayes achieves moderate $\Omega$ scores with the EfficientNet-B0, MobileNet-v3 (Large), and EfficientNet-B1 backbones and a poor score with the ResNet-18 backbone. This could be because the features learned with ResNet-18 are not independent, which is an assumption of the Naive Bayes method. Unsurprisingly, all methods achieve the worst $\Omega$ scores with the ResNet-18 architecture, which is because it is the largest network and requires the most memory and compute.

\subsubsection{Backbone CNN Comparisons}

Moreover, we were interested in seeing how performance varied for each online continual learning method across CNN architectures. To study this, we show the final top-1 accuracy scores achieved by each method when trained using the full instance ordering in Table~\ref{tab:open-loris-full-results}. For nearly all online continual learners (with the exception of NCM), performance using the MobileNet-v3 (Small) and ResNet-18 backbones is the worst, while performance using the EfficientNet-B0 and EfficientNet-B1 backbones is the best. This is interesting as many existing continual learning methods have focused on pairing new algorithmic components for overcoming catastrophic forgetting with the ResNet-18 architecture~\citep{rebuffi2016icarl,hou2019unified,wu2019large,hayes2019remind}. Since performance is better using EfficientNets and they require less memory and compute than ResNet-18, we urge future researchers to consider alternative architectures for studying continual learning.

\begin{table}[t]
\caption{Final accuracy (\%) values for each online continual learner using the \emph{instance} ordering of the OpenLORIS dataset. 
We also report the mean accuracy (\%) aggregated across CNN architectures.
We format the \underline{\textbf{first}}, \textbf{second} and \underline{third} best overall results. \label{tab:open-loris-full-results}}
\centering
\begin{tabular}{lcccccc}
\toprule
\textsc{Method} & \textsc{MNet-S} & \textsc{MNet-L} & \textsc{ENet-B0} & \textsc{ENet-B1} & \textsc{RN-18} & \textsc{Mean} \\ 
\midrule
Perceptron & 79.3 & 88.0 & 93.5 & 94.2 & 79.6 & 86.9 \\
Fine-Tune & 83.5 & 91.5 & 95.8 & 96.3 & 82.1 & 89.8 \\
Naive Bayes & 31.1 & 52.6 & 78.0 & 78.8 & 1.5 & 48.4 \\ 
SOvR & 37.4 & 47.7 & 73.9 & 72.4 & 34.6 & 53.2 \\
NCM & 72.9 & 78.9 & 85.9 & 86.7 & 79.7 & 80.8 \\
Replay (2pc) & 89.3 & 94.2 & 97.0 & 97.4 & 90.7 & \underline{93.7} \\
Replay (20pc) & 92.1 & 95.6 & 97.7 & 97.8 & 92.9 & \textbf{95.2} \\
SLDA & 95.6 & 98.2 & 98.8 & 98.8 & 95.0 & \underline{\textbf{97.3}} \\
\bottomrule
\end{tabular}
\end{table}

\subsection{Overall Results on Places-365 and Places-LT}

Next, we looked at which continual learning method was the most effective, regardless of the CNN architecture, on both variants of the Places dataset. Average top-1 accuracies on Places-365 and Places-LT for all continual leaning methods averaged across CNN architectures are given in Table~\ref{tab:overall-results}, where the overall accuracy is computed as the harmonic mean of the class-iid and iid runs to emphasize that it is important to do well on both extremes. The raw top-1 accuracy values with each CNN backbone are in supplemental materials. SLDA performed best overall, followed by replay (20pc) and NCM. As expected, in the iid setting, fine-tune performed well and in the class-iid setting it suffered from severe catastrophic forgetting. Similarly, the perceptron performed well on the iid ordering, but struggled with the class-iid ordering. This is likely because when the perceptron misclassifies an example, it updates the weight for the correct class and the weight for the highest predicted incorrect class. Since classes are not revisited in the class-iid setting, it is likely that the perceptron misclassifies new examples as previous classes and updates the previous class weight vectors. This leads to perturbations in the previous weights, causing catastrophic forgetting. While Naive Bayes and SOvR were unaffected by data ordering, their performance was much worse than the top-performing methods. For Naive Bayes, this is likely because its feature independence assumptions are violated.

Since Places-365 and Places-LT use the same test set, but different training sets, we can directly compare the performance of online continual learners on the two datasets to see how robust they are to dataset imbalance. Interestingly, the NCM method only exhibits a 3.1\% loss in performance when trained on the long-tailed dataset, while the SLDA and replay (20pc) methods have larger gaps of 7.8\% and 7.0\%, respectively. This indicates that the NCM method is more robust to dataset imbalance than SLDA or replay (20pc). This is likely because SLDA assumes equal class covariances, which could be violated in the imbalanced regime. Similarly, performance of the replay method could potentially be improved by selectively replaying more examples from underrepresented classes. While most methods exhibit worse performance when trained on the long-tailed dataset, the perceptron and fine-tune methods exhibit minor performance gains, but still perform poorly overall. While we randomly shuffled classes for the Places-365 and Places-LT datasets, it could be an interesting future study to understand how the order of classes impacts performance on long-tailed datasets (e.g., ordering classes based on their number of training examples).

\section{Discussion and Conclusion}
\label{sec:recommendations-discussion}

Here we described a real-world problem where continual learning is needed: learning on embedded devices. From this need, we are able to specify the essential capabilities for these methods. In general, many existing continual learning frameworks do not meet the needs of embedded devices since they require processing data in large batches, require task labels to be provided during inference, or cannot learn from data streams presented in any order. To address the needs of continual learning for embedded devices, we compared seven online continual learning methods when paired with backbone CNN architectures designed for mobile and embedded applications to identify which methods fit our proposed criteria. We conducted experiments on three high-resolution image classification datasets to evaluate the robustness of the learners to scale (Places-365), imbalanced data streams (Places-LT), and realistic video data streams of specific object instances filmed under various domain settings (OpenLORIS). We compared learners across three axes: classification efficacy, memory, and compute. Overall, we found that the replay (20pc) and SLDA models achieve strong performance on both orderings of the OpenLORIS dataset. However, when memory and compute time are factored in, the NCM method strikes the best trade-off between classification efficacy and efficiency on the OpenLORIS dataset. We also found that methods performed consistently the best when paired with an EfficientNet CNN. Finally, we found that the SLDA, replay (20pc), and NCM methods performed best on both variants and orderings of the Places dataset. We urge future embedded continual learning researchers to consider using the NCM, SLDA, and replay (20pc) models as baselines as they achieve strong classification efficacy while also minimizing memory and compute.

\begin{table}[t]
\caption{Final accuracy (\%) summary statistics aggregated across CNN architectures to compare online continual learning methods. The iid and class-iid results are computed with the arithmetic mean across CNN architectures and the overall performance is computed as the harmonic mean (H-Mean) of these two numbers. An ideal method would achieve strong results regardless of ordering. 
We format the \underline{\textbf{first}}, \textbf{second} and \underline{third} best overall results.
\label{tab:overall-results}}
\centering

\begin{tabular}{lcccccc}
\toprule
       & \multicolumn{3}{c}{\textsc{Places-365}}                                                        & \multicolumn{3}{c}{\textsc{Places-LT}}                                                         \\
       \cmidrule(r){2-4} \cmidrule(r){5-7}
\textsc{Method} & \multicolumn{1}{c}{\textsc{iid}} & \multicolumn{1}{c}{\textsc{class-iid}} & \multicolumn{1}{c}{\textsc{H-Mean}} & \multicolumn{1}{c}{\textsc{iid}} & \multicolumn{1}{c}{\textsc{class-iid}} & \multicolumn{1}{c}{\textsc{H-Mean}} \\
\midrule
    Perceptron & 32.2 & 0.9 & 1.8 & 18.0 & 4.1 & 6.7 \\
    Fine-Tune & 44.0 & 2.9 & 5.4 & 20.4 & 5.0 & 8.1 \\
    Naive Bayes & 12.7 & 12.7 & 12.7 & 9.6 & 9.6 & 9.6 \\
    SOvR & 20.0 & 20.0 & 20.0 & 17.8 & 17.8 & 17.8 \\
    NCM & 33.9 & 33.9 & \underline{33.9} & 30.8 & 30.8 & \textbf{30.8} \\
    Replay (2pc) & 43.4 & 20.7 & 28.1 & 29.7 & 20.6 & 24.3 \\
    Replay (20pc) & 44.1 & 32.3 & \textbf{37.3} & 31.4 & 29.2 & \underline{30.3} \\
    SLDA & 39.3 & 39.3 & \underline{\textbf{39.3}} & 31.5 & 31.5 & \underline{\textbf{31.5}} \\
    \bottomrule
\end{tabular}
\end{table}

While we investigated the practicality of several online continual learners for embedded applications, there are several future research directions to explore. First, we focused on studying online learners that use fixed features from CNNs trained using supervised learning on ImageNet-1k. Recently, features learned using self-supervised learning have demonstrated strong performance on many downstream tasks~\citep{chen2020simple,chen2020big,grill2020bootstrap,caron2020unsupervised,gallardo2021selfsupervised} and it would be interesting to apply them to embedded devices. The challenge is that most self-supervised techniques have been designed for large CNNs (e.g., ResNet-50) and perform poorly when applied directly to mobile architectures~\citep{fang2021seed,abbasi2020compress}. Moreover, we extracted pre-trained features from only the penultimate layer of each CNN. It could be interesting to explore the use of features from additional layers of the CNN as well. Moreover, the ImageNet dataset has its own limitations and might not yield the best features for every application, so it would be interesting to explore additional pre-training datasets.

Alternatively, methods that update feature representations could be considered in the future. There are two main challenges with continually updating representations for embedded applications: updating more features requires more memory and compute and updating feature representations can result in concept drift. Beyond this, improvements could be made to algorithms to better handle out-of-domain generalization and low-shot learning. While we found some methods performed better on the low-shot instance order than others, performance across all methods was worse than training on all object instances indicating the need for more robust low-shot learners. Moreover, we focused on supervised image classification, but additional capabilities such as object detection or segmentation could be studied for embedded devices in the future. For example, object detection requires both image classification and regression. The online classification methods explored in this paper could be combined with streaming regression models to perform lightweight, online object detection. Beyond this, we focused on CNN architectures  designed for mobile and embedded applications; however, alternative methods to improve efficiency could be considered (e.g., pruning, quantization, etc.).

We believe that implementing supervised online continual learning algorithms on embedded devices requires the general criteria outlined in this paper. We established baselines for online continual learning methods to better understand which algorithms work best under a variety of settings that are useful for embedded applications (e.g., imbalanced data streams, large-scale data streams, video streams, and low-shot video streams). The insights learned from these experiments can provide future researchers with a starting point for specific online embedded applications such as AR/VR headsets, smartphones, robots, smart toys, and more. Performing embedded continual learning can potentially reduce latency, power consumption, privacy concerns, and the overall carbon footprint. Some challenges that will need to be considered include the efficient implementation of algorithms on specialized hardware and the background collection and processing of data that is provided to the continual learners.


\subsubsection*{Acknowledgments}
This work was supported in part by NSF awards \#1909696, \#2047556, and \#DGE-2125362. The views and conclusions contained herein are those of the authors and should not be interpreted as representing the official policies or endorsements of any sponsor. We thank James Arnold and Robik Shrestha for their comments and useful discussions.

\bibliography{collas2022_conference}
\bibliographystyle{collas2022_conference}

\clearpage
\appendix
\section{Appendix}

\subsection{Implementation Details}
\label{sec:implementation}

We use the backbone CNN implementations and supervised ImageNet-1k pre-trained checkpoints for MobileNet-v3, EfficientNet, and ResNet-18 from torchvision. For the fine-tune and replay models, we select the best learning rate from \{0.1, 0.01, 0.001, 0.0001\}. For Places-365, the best learning rate is 0.0001. For Places-LT and OpenLORIS, the best learning rate is 0.001. For fine-tune and replay, we use a weight decay factor of $10^{-5}$, a momentum of 0.9, and the stochastic gradient descent optimizer. For replay, we follow \citet{gallardo2021selfsupervised} and randomly select 50 samples from the replay buffer to combine with the new sample to update the model. For replay, we study two buffer sizes: 1) storing 20 examples per class, which is common in continual learning literature~\citep{rebuffi2016icarl,wu2019large} and 2) storing 2 examples per class, which requires a similar amount of memory to the other online continual learners studied in this paper. For SLDA and Naive Bayes, we follow \citet{hayes2019lifelong} and use a shrinkage value of $10^{-4}$.

\subsection{Additional Results}

\subsubsection{Places-365 and Places-LT}

\begin{table}[h]
\caption{Final accuracy (\%) results on the full Places-365 dataset with the iid and class-iid data orderings.  \label{tab:places-results}}
\centering
\resizebox{\textwidth}{!}{%
\begin{tabular}{lcccccccccc}
\toprule
& \multicolumn{5}{c}{\textsc{iid}} & \multicolumn{5}{c}{\textsc{class-iid}} \\
\cmidrule(r){2-6} \cmidrule(r){7-11}
\textsc{Method} & \textsc{MNet-S} & \textsc{MNet-L} & \textsc{ENet-B0} & \textsc{ENet-B1} & \textsc{RN-18} & \textsc{MNet-S} & \textsc{MNet-L} & \textsc{ENet-B0} & \textsc{ENet-B1} & \textsc{RN-18}  \\ 
\midrule
Perceptron & 29.4 & 33.7 & 34.7 & 33.4 & 29.6 & 0.5 & 0.6 & 1.7 & 1.6 & 0.3 \\
Fine-Tune & 41.2 & 45.3 & 46.6 & 45.9 & 40.8 & 0.7 & 0.9 & 6.1 & 6.3 & 0.4 \\
Naive Bayes & 3.0 & 9.4 & 25.5 & 25.2 & 0.3 & 3.0 & 9.4 & 25.5 & 25.2 & 0.3 \\
SOvR & 9.6 & 16.9 & 29.1 & 28.3 & 16.0 & 9.6 & 16.9 & 29.1 & 28.3 & 16.0 \\
NCM & 29.4 & 34.0 & 36.8 & 36.4 & 32.8 & 29.4 & 34.0 & 36.8 & 36.4 & 32.8 \\
Replay (2pc) & 41.0 & 44.8 & 45.6 & 45.1 & 40.4 & 16.2 & 19.3 & 26.2 & 25.6 & 16.4 \\
Replay (20pc) & 41.3 & 45.7 & 46.3 & 45.7 & 41.5 & 29.5 & 32.6 & 35.8 & 35.0 & 28.8 \\
SLDA & 36.9 & 40.3 & 41.7 & 41.0 & 36.6 & 36.9 & 40.3 & 41.7 & 41.0 & 36.6 \\
\bottomrule
\end{tabular}
}
\end{table}

\begin{table}[h]
\caption{Final accuracy (\%) results on the long-tailed Places-LT dataset with the iid and class-iid data orderings. Each result is the average over 3 runs with different permutations of the data. \label{tab:places-lt-results}}
\centering
\resizebox{\textwidth}{!}{%
\begin{tabular}{lcccccccccc}
\toprule
& \multicolumn{5}{c}{\textsc{iid}} & \multicolumn{5}{c}{\textsc{class-iid}} \\
\cmidrule(r){2-6} \cmidrule(r){7-11}
\textsc{Method} & \textsc{MNet-S} & \textsc{MNet-L} & \textsc{ENet-B0} & \textsc{ENet-B1} & \textsc{RN-18} & \textsc{MNet-S} & \textsc{MNet-L} & \textsc{ENet-B0} & \textsc{ENet-B1} & \textsc{RN-18}  \\ 
\midrule
Perceptron & 15.2 & 18.5 & 21.3 & 20.6 & 14.5 & 1.7 & 2.8 & 7.1 & 7.3 & 1.5 \\
Fine-Tune & 16.9 & 21.0 & 23.8 & 23.0 & 17.3 & 1.8 & 3.4 & 9.7 & 9.7 & 0.6 \\
Naive Bayes & 1.5 & 5.0 & 19.9 & 21.3 & 0.1 & 1.5 & 5.0 & 19.9 & 21.3 & 0.1 \\
SOvR & 8.9 & 14.9 & 26.2 & 24.5 & 14.6 & 8.9 & 14.9 & 26.2 & 24.5 & 14.6 \\
NCM & 26.5 & 31.0 & 33.6 & 32.9 & 30.0 & 26.5 & 31.0 & 33.6 & 32.9 & 30.0 \\
Replay (2pc) & 27.3 & 29.9 & 32.6 & 31.9 & 26.7 & 16.6 & 20.5 & 24.1 & 23.9 & 17.9 \\
Replay (20pc) & 29.4 & 32.2 & 34.0 & 33.1 & 28.5 & 26.6 & 29.5 & 31.9 & 31.3 & 26.8 \\
SLDA & 29.0 & 31.8 & 33.8 & 32.8 & 30.0 & 29.0 & 31.9 & 33.8 & 32.8 & 30.0 \\
\bottomrule
\end{tabular}
}
\end{table}

We include the final top-1 accuracies achieved by each online continual learning method with each CNN backbone on the Places-365 and Places-LT datasets in Table~\ref{tab:places-results} and Table~\ref{tab:places-lt-results}, respectively. These results are aggregated in Table~\ref{tab:overall-results}.

\subsubsection{OpenLORIS}

\begin{figure*}[t]
    \centering
    \begin{subfigure}[t]{0.48\linewidth}
        \includegraphics[width=\linewidth]{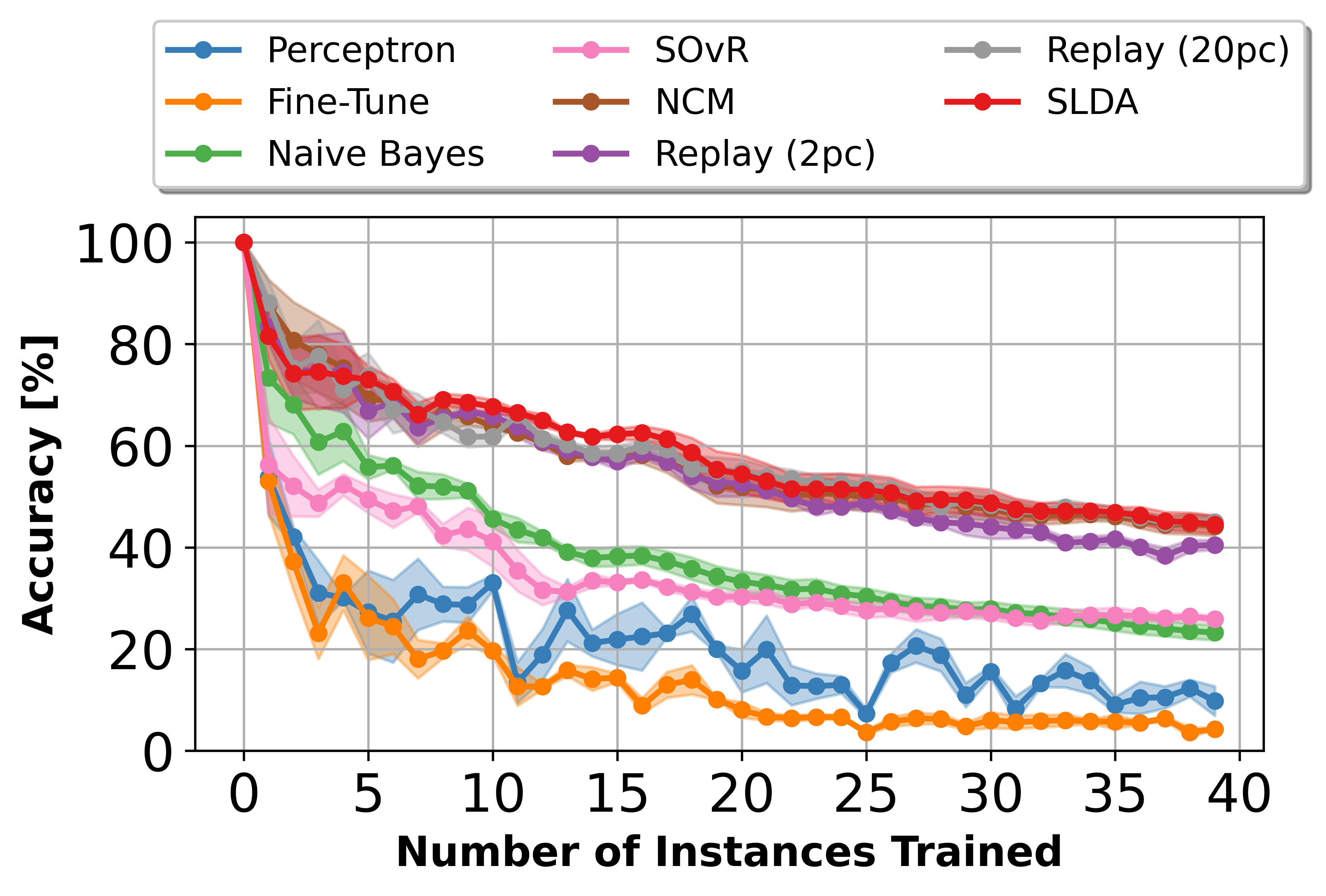}
        \caption{MobileNet-v3 (Small)}
        \label{fig:low-shot-mb-small}
    \end{subfigure} %
    \centering
    \begin{subfigure}[t]{0.48\linewidth}
        \includegraphics[width=\linewidth]{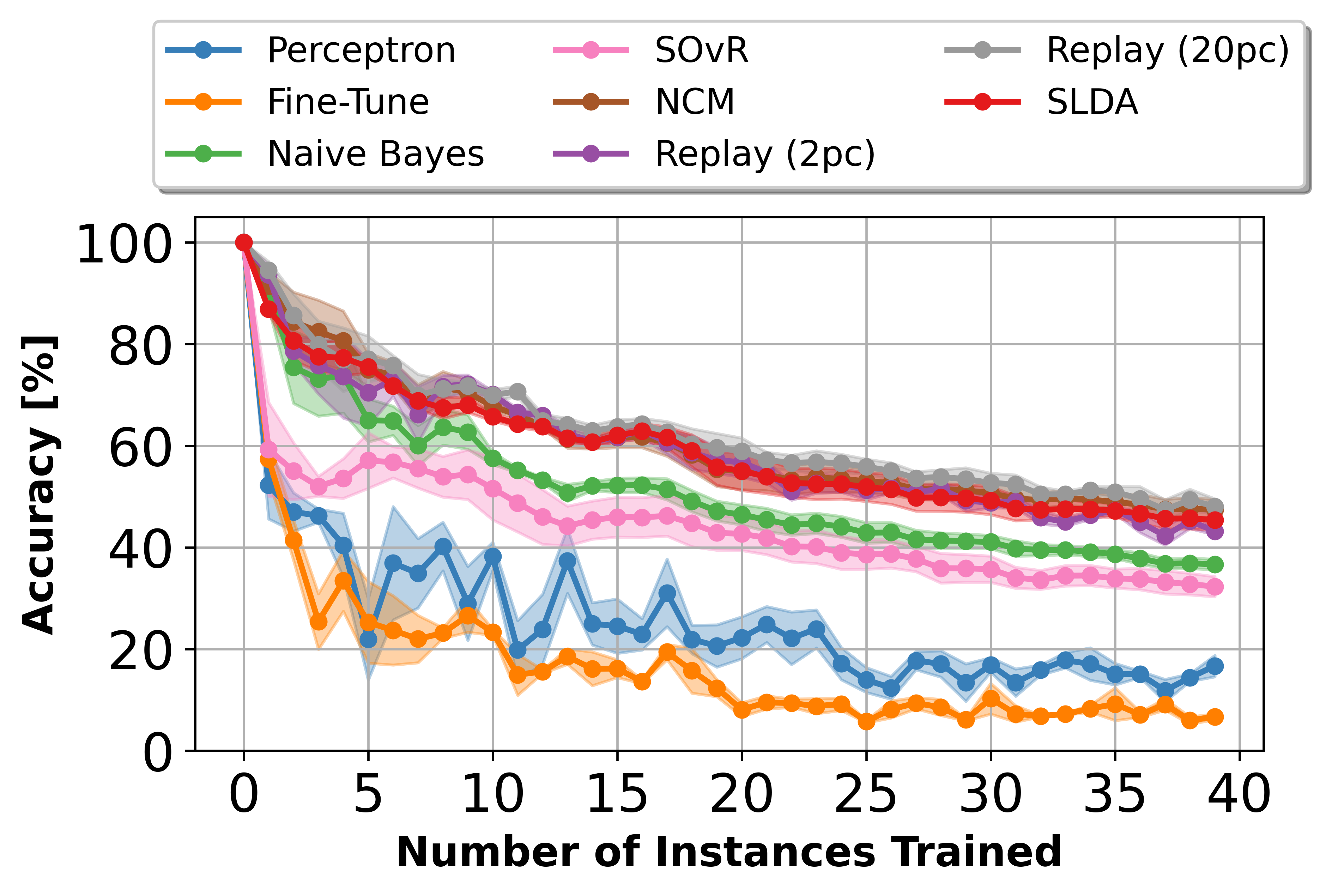}
        \caption{MobileNet-v3 (Large)}
        \label{fig:low-shot-mb-large}
    \end{subfigure} %
    \\
    \centering
    \begin{subfigure}[t]{0.48\linewidth}
        \includegraphics[width=\linewidth]{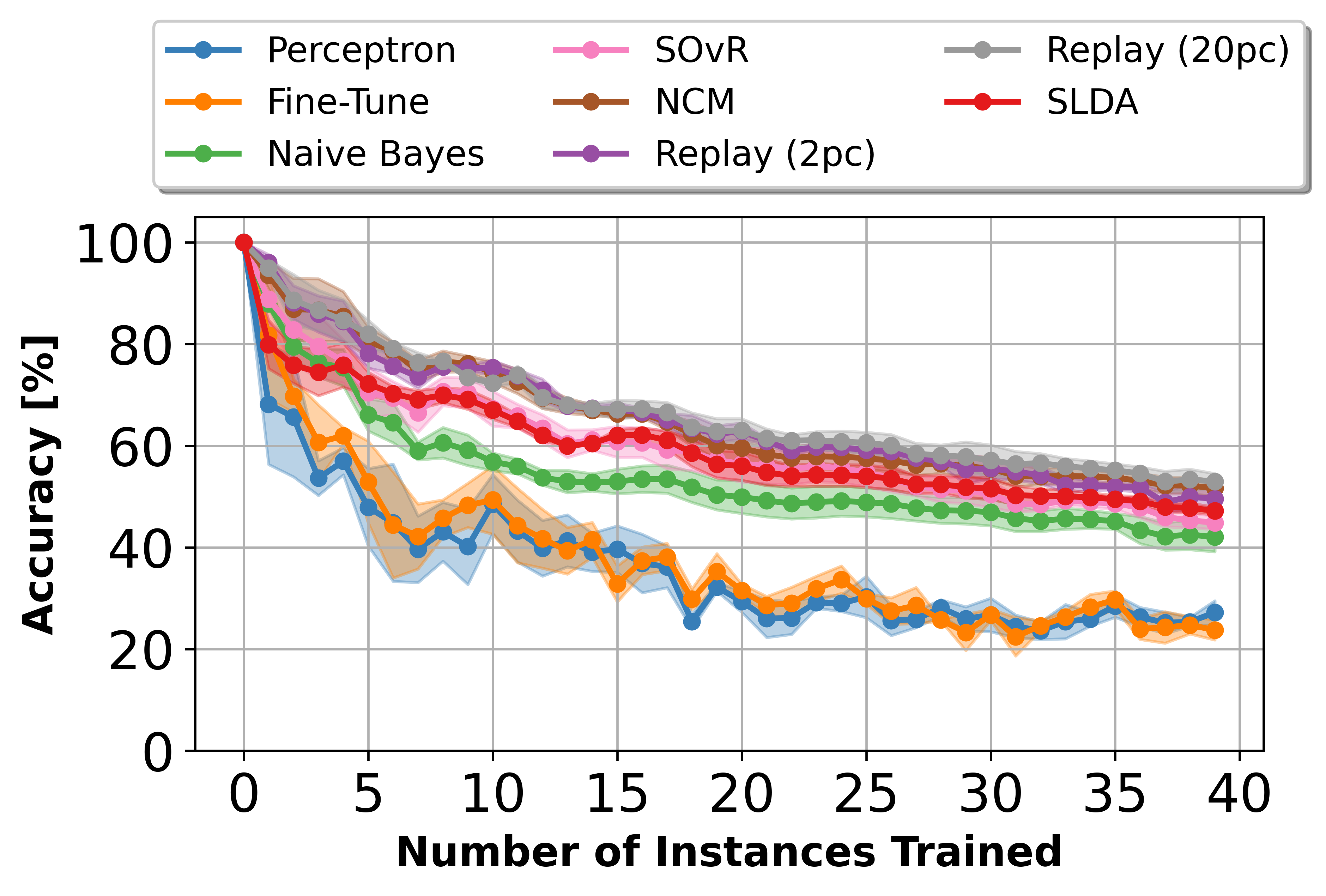}
        \caption{EfficientNet-B0}
        \label{fig:low-shot-en-b0}
    \end{subfigure} %
    \centering
    \begin{subfigure}[t]{0.48\linewidth}
        \includegraphics[width=\linewidth]{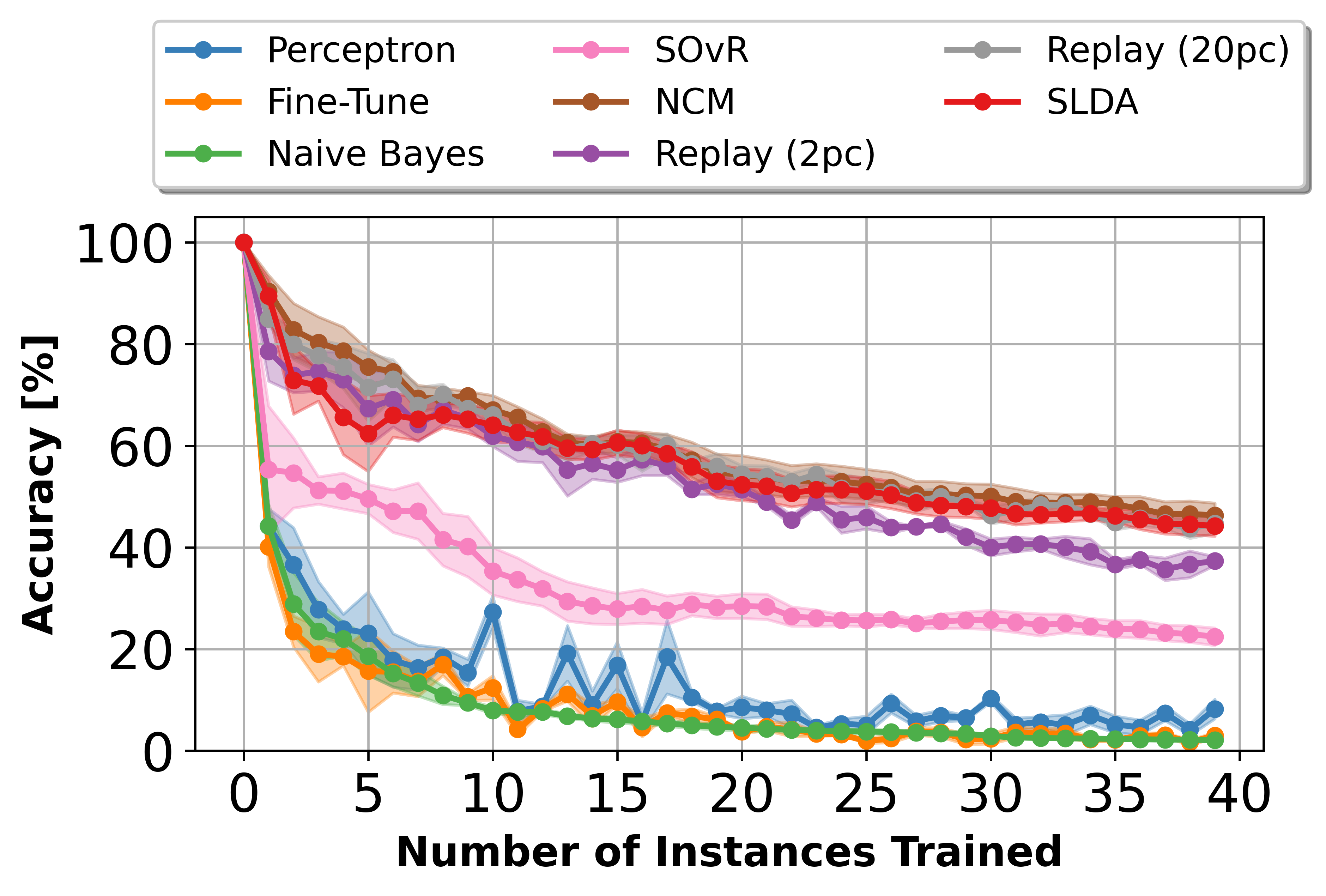}
        \caption{ResNet-18}
        \label{fig:low-shot-rn-18}
    \end{subfigure} %
    \caption{Learning curves of each online continual learner with various backbone CNNs when evaluated using the low-shot instance ordering of the OpenLORIS dataset. Each plot is the average over 3 runs with different instance permutations. The standard error over runs is denoted by the shaded regions.
    }
    \label{fig:low-shot-instance-plots}
\end{figure*}
\begin{table}[t]
\caption{Final accuracy (\%) values for each online continual learner using the \emph{low-shot instance} ordering of the OpenLORIS dataset. 
We also report the mean accuracy (\%) aggregated across CNN architectures.
We format the \underline{\textbf{first}}, \textbf{second} and \underline{third} best overall results. \label{tab:open-loris-low-shot-results}}
\centering
\begin{tabular}{lcccccc}
\toprule
\textsc{Method} & \textsc{MNet-S} & \textsc{MNet-L} & \textsc{ENet-B0} & \textsc{ENet-B1} & \textsc{RN-18} & \textsc{Mean} \\ 
\midrule
Perceptron & 9.8 & 16.7 & 27.2 & 28.3 & 8.2 & 18.0 \\
Fine-Tune & 4.3 & 6.7 & 23.8 & 23.6 & 3.0 & 12.3 \\
Naive Bayes & 23.3 & 36.7 & 42.1 & 39.9 & 2.1 & 28.8 \\
SOvR & 25.9 & 32.3 & 44.9 & 45.9 & 22.4 & 34.3 \\
NCM & 44.2 & 47.4 & 51.6 & 51.4 & 46.3 & \textbf{48.2} \\
Replay (2pc) & 40.5 & 43.2 & 49.6 & 49.1 & 37.3 & 43.9 \\
Replay (20pc) & 45.0 & 48.1 & 53.0 & 53.0 & 44.7 & \underline{\textbf{48.8}} \\
SLDA & 44.5 & 45.4 & 47.2 & 46.0 & 44.2 & \underline{45.5} \\
\bottomrule
\end{tabular}
\end{table}

In Fig.~\ref{fig:low-shot-instance-plots}, we show learning curves for each online continual learning method when trained with the low-shot instance ordering of the OpenLORIS dataset for various CNN backbones. Learning curves using the EfficientNet-B1 backbone are in Fig.~\ref{fig:open-loris-learning-curve}. While results vary slightly across CNN architectures, the replay (20pc) and NCM methods are consistently the top performers. This indicates that NCM and replay (20pc) generalize the best to out-of-domain inputs.

We show the final top-1 accuracy scores achieved by each method when trained using the low-shot instance ordering in Table~\ref{tab:open-loris-low-shot-results}. These scores coincide with the final accuracy values in Fig.~\ref{fig:open-loris-learning-curve} and Fig.~\ref{fig:low-shot-instance-plots}.



To compute the NetScore values in Table~\ref{tab:open-loris-summarized-results}, we take into account a learner's classification efficacy (final accuracy), the total number of parameters required to store the backbone and continual learning model, and the compute time required to run the experiment in seconds. To examine how each of these factors contribute directly to a learner's NetScore, we show the raw efficacy, memory, and compute for each learner with each backbone in Tables~\ref{tab:eff-mem-comp-mnsmall}-\ref{tab:eff-mem-comp-rn18}.

\begin{table}[h]
\caption{Classification efficacy (final accuracy), memory (number of parameters), compute (experiment run-time in seconds), and associated NetScore ($\Omega$) values for each online continual learner using the \emph{low-shot instance} ordering of the OpenLORIS dataset with the \textbf{MobileNet-v3 (Small)} backbone CNN. We format the \underline{\textbf{first}}, \textbf{second} and \underline{third} best overall results. \label{tab:eff-mem-comp-mnsmall}}
\centering
\begin{tabular}{lcccc}
\toprule
\textsc{Method} & \textsc{Efficacy} & \textsc{Memory} & \textsc{Compute} & \textsc{$\Omega$} \\ 
\midrule
Perceptron & 9.8 & 950048 & 1041 & -12.3 \\
Fine-Tune & 4.3 & 950048 & 1040 & -45.5 \\
Naive Bayes & 23.2 & 996128 & 1260 & 21.1 \\
SOvR & 25.9 & 950048 & 1414 & 25.0 \\
NCM & 44.2 & 950048 & 1035 & \underline{\textbf{48.0}} \\
Replay (2pc) & 40.5 & 996128 & 1053 & 44.2 \\
Replay (20pc) & 45.0 & 1410848 & 1052 & \underline{46.7} \\
SLDA & 44.5 & 1281824 & 1040 & \textbf{46.8} \\
\bottomrule
\end{tabular}
\end{table}
\begin{table}[h]
\caption{Classification efficacy (final accuracy), memory (number of parameters), compute (experiment run-time in seconds), and associated NetScore ($\Omega$) values for each online continual learner using the \emph{low-shot instance} ordering of the OpenLORIS dataset with the \textbf{MobileNet-v3 (Large)} backbone CNN. We format the \underline{\textbf{first}}, \textbf{second} and \underline{third} best overall results. \label{tab:eff-mem-comp-mnlarge}}
\centering
\begin{tabular}{lcccc}
\toprule
\textsc{Method} & \textsc{Efficacy} & \textsc{Memory} & \textsc{Compute} & \textsc{$\Omega$} \\ 
\midrule
Perceptron & 16.7 & 3010352 & 1082 & 3.0 \\
Fine-Tune & 6.7 & 3010352 & 1084 & -33.4 \\
Naive Bayes & 36.6 & 3087152 & 1329 & 33.4 \\
SOvR & 32.3 & 3010352 & 1583 & 27.6 \\
NCM & 47.4 & 3010352 & 1078 & \underline{\textbf{44.8}} \\
Replay (2pc) & 43.2 & 3087152 & 1093 & 40.9 \\
Replay (20pc) & 48.1 & 3778352 & 1094 & \textbf{44.2} \\
SLDA & 45.4 & 3931952 & 1082 & \underline{41.8} \\
\bottomrule
\end{tabular}
\end{table}
\begin{table}[h]
\caption{Classification efficacy (final accuracy), memory (number of parameters), compute (experiment run-time in seconds), and associated NetScore ($\Omega$) values for each online continual learner using the \emph{low-shot instance} ordering of the OpenLORIS dataset with the \textbf{EfficientNet-B0} backbone CNN. We format the \underline{\textbf{first}}, \textbf{second} and \underline{third} best overall results. \label{tab:eff-mem-comp-enb0}}
\centering
\begin{tabular}{lcccc}
\toprule
\textsc{Method} & \textsc{Efficacy} & \textsc{Memory} & \textsc{Compute} & \textsc{$\Omega$} \\ 
\midrule
Perceptron & 27.2 & 4058748 & 1204 & 20.6 \\
Fine-Tune & 23.8 & 4058748 & 1198 & 15.2 \\
Naive Bayes & 42.1 & 4161148 & 1475 & 36.9 \\
SOvR & 44.9 & 4058748 & 1635 & 39.1 \\
NCM & 51.6 & 4058748 & 1196 & \underline{\textbf{46.2}} \\
Replay (2pc) & 49.6 & 4161148 & 1211 & \underline{44.4} \\
Replay (20pc) & 53.0 & 5082748 & 1211 & \textbf{46.1} \\
SLDA & 47.2 & 5697148 & 1202 & 40.9 \\
\bottomrule
\end{tabular}
\end{table}
\begin{table}[h]
\caption{Classification efficacy (final accuracy), memory (number of parameters), compute (experiment run-time in seconds), and associated NetScore ($\Omega$) values for each online continual learner using the \emph{low-shot instance} ordering of the OpenLORIS dataset with the \textbf{EfficientNet-B1} backbone CNN. We format the \underline{\textbf{first}}, \textbf{second} and \underline{third} best overall results. \label{tab:eff-mem-comp-enb1}}
\centering
\begin{tabular}{lcccc}
\toprule
\textsc{Method} & \textsc{Efficacy} & \textsc{Memory} & \textsc{Compute} & \textsc{$\Omega$} \\ 
\midrule
Perceptron & 28.3 & 6564384 & 1511 & 18.6 \\
Fine-Tune & 23.6 & 6564384 & 1502 & 11.4 \\
Naive Bayes & 39.9 & 6666784 & 1719 & 31.6 \\
SOvR & 45.9 & 6564384 & 1777 & 37.2 \\
NCM & 51.4 & 6564384 & 1501 & \textbf{42.5} \\
Replay (2pc) & 49.1 & 6666784 & 1518 & \underline{40.6} \\
Replay (20pc) & 53.0 & 7588384 & 1513 & \underline{\textbf{43.0}} \\
SLDA & 46.0 & 8202784 & 1509 & 36.9 \\
\bottomrule
\end{tabular}
\end{table}
\begin{table}[h]
\caption{Classification efficacy (final accuracy), memory (number of parameters), compute (experiment run-time in seconds), and associated NetScore ($\Omega$) values for each online continual learner using the \emph{low-shot instance} ordering of the OpenLORIS dataset with the \textbf{ResNet-18} backbone CNN. We format the \underline{\textbf{first}}, \textbf{second} and \underline{third} best overall results. \label{tab:eff-mem-comp-rn18}}
\centering
\begin{tabular}{lcccc}
\toprule
\textsc{Method} & \textsc{Efficacy} & \textsc{Memory} & \textsc{Compute} & \textsc{$\Omega$} \\ 
\midrule
Perceptron & 8.2 & 11196992 & 1076 & -31.8 \\
Fine-Tune & 3.0 & 11196992 & 1072 & -71.8 \\
Naive Bayes & 2.1 & 11237952 & 1314 & -87.4 \\
SOvR & 22.4 & 11196992 & 1567 & 6.5 \\
NCM & 46.3 & 11196992 & 1073 & \underline{\textbf{37.4}} \\
Replay (2pc) & 37.3 & 11237952 & 1089 & 28.7 \\
Replay (20pc) & 44.7 & 11606592 & 1083 & \textbf{35.7} \\
SLDA & 44.2 & 11459136 & 1074 & \underline{35.4} \\
\bottomrule
\end{tabular}
\end{table}

\begin{table}[t]
\caption{$\Omega$ NetScore values for each online continual learner using the \emph{low-shot instance} ordering of the OpenLORIS dataset. Here, we use $\alpha=2$, $\beta=\gamma=0.5$ to compute the NetScore as originally done in \citet{wong2019netscore}.
We also report the mean accuracy (\%) aggregated across CNN architectures.
We format the \underline{\textbf{first}}, \textbf{second} and \underline{third} best overall results.
\label{tab:open-loris-summarized-results-orig-params}}
\centering
\begin{tabular}{lcccccc}
\toprule
\textsc{Method} & \textsc{MNet-S} & \textsc{MNet-L} & \textsc{ENet-B0} & \textsc{ENet-B1} & \textsc{RN-18} & \textsc{Mean} \\ 
\midrule
Perceptron & -115.9 & -106.5 & -91.0 & -96.5 & -147.8 & -111.5 \\
Fine-Tune & -149.0 & -142.9 & -96.3 & -103.7 & -187.8 & -135.9 \\
Naive Bayes & -83.7 & -77.3 & -75.8 & -84.2 & -204.5 & -105.1 \\
SOvR & -80.0 & -83.9 & -74.0 & -78.7 & -111.4 & -85.6 \\
NCM & -55.5 & -64.7 & -65.3 & -72.5 & -78.7 & \underline{\textbf{-67.3}} \\
Replay (2pc) & -59.7 & -68.8 & -67.3 & -74.6 & -87.5 & \underline{-71.6} \\
Replay (20pc) & -58.9 & -66.5 & -66.6 & -72.8 & -80.5 & \textbf{-69.1} \\
SLDA & -58.3 & -69.1 & -72.3 & -79.3 & -80.8 & -72.0 \\
\bottomrule
\end{tabular}
\end{table}

For completeness, Table~\ref{tab:open-loris-summarized-results-orig-params} contains NetScore values for each continual learner when computed using the original parameters suggested by \citet{wong2019netscore}, i.e., $\alpha=2$, $\beta=\gamma=0.5$. The NetScore values in Table~\ref{tab:open-loris-summarized-results-orig-params} follow similar trends as the NetScore values using $\alpha=2$, $\beta=\gamma=0.25$ from Table~\ref{tab:open-loris-summarized-results}. That is, the best performing model is NCM, followed by replay (20pc).

\begin{figure*}[t]
    \centering
    \begin{subfigure}[t]{0.32\linewidth}
        \includegraphics[width=\linewidth]{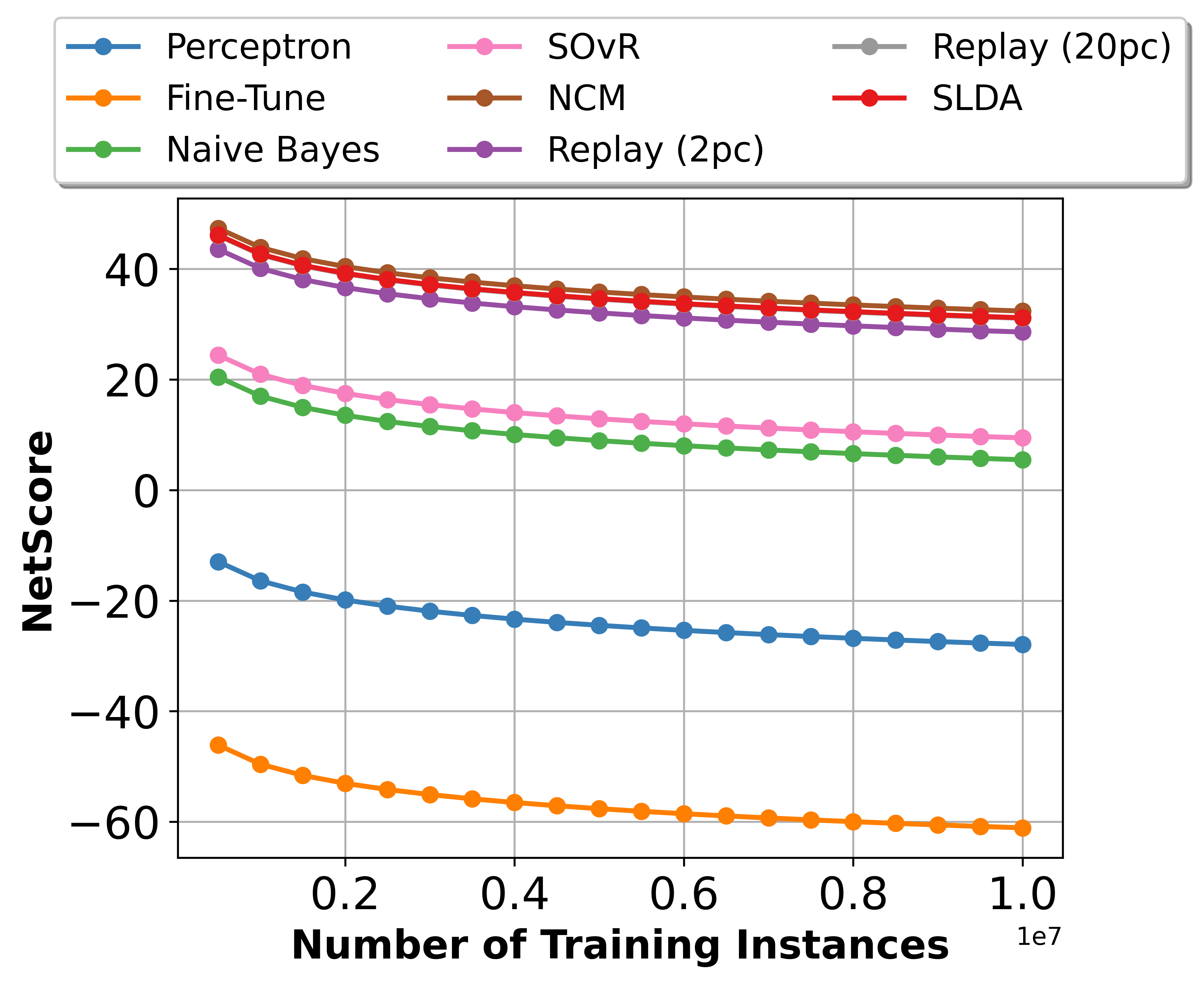}
        \caption{MobileNet-v3 (Small)}
    \end{subfigure} %
    \centering
    \begin{subfigure}[t]{0.32\linewidth}
        \includegraphics[width=\linewidth]{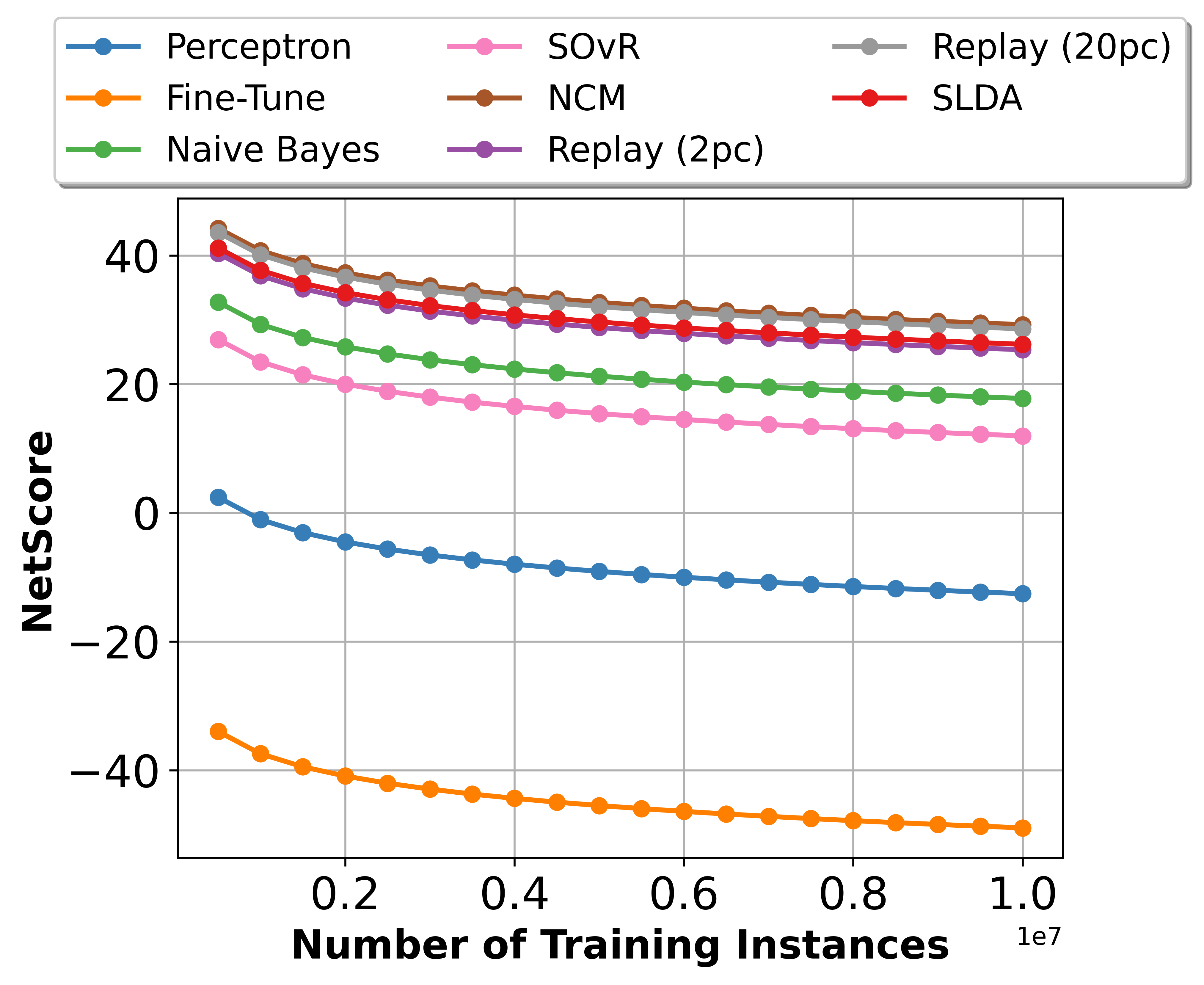}
        \caption{MobileNet-v3 (Large)}
    \end{subfigure} %
    \centering
    \begin{subfigure}[t]{0.32\linewidth}
        \includegraphics[width=\linewidth]{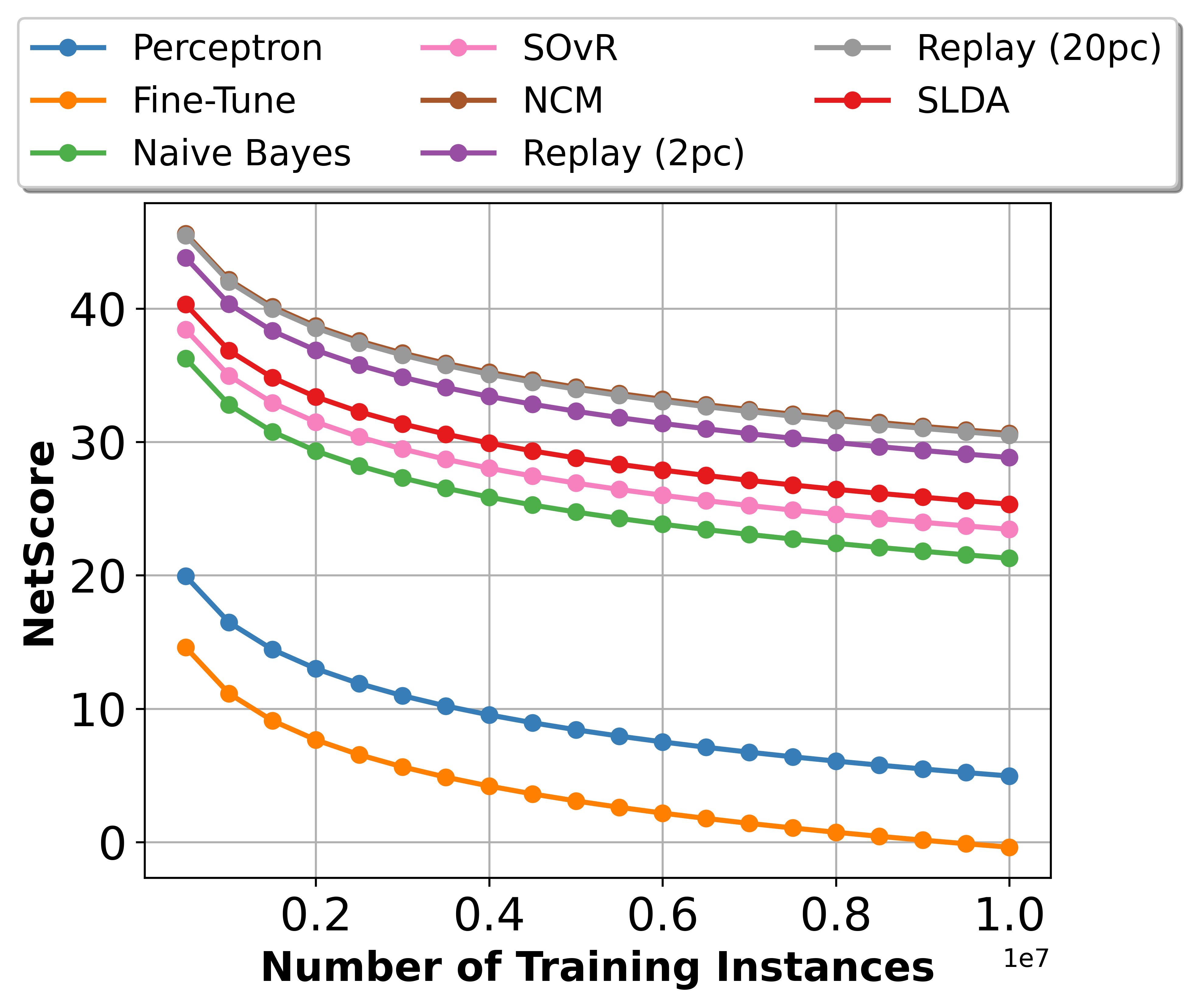}
        \caption{EfficientNet-B0}
    \end{subfigure} %
    \\
    \centering
    \begin{subfigure}[t]{0.32\linewidth}
        \includegraphics[width=\linewidth]{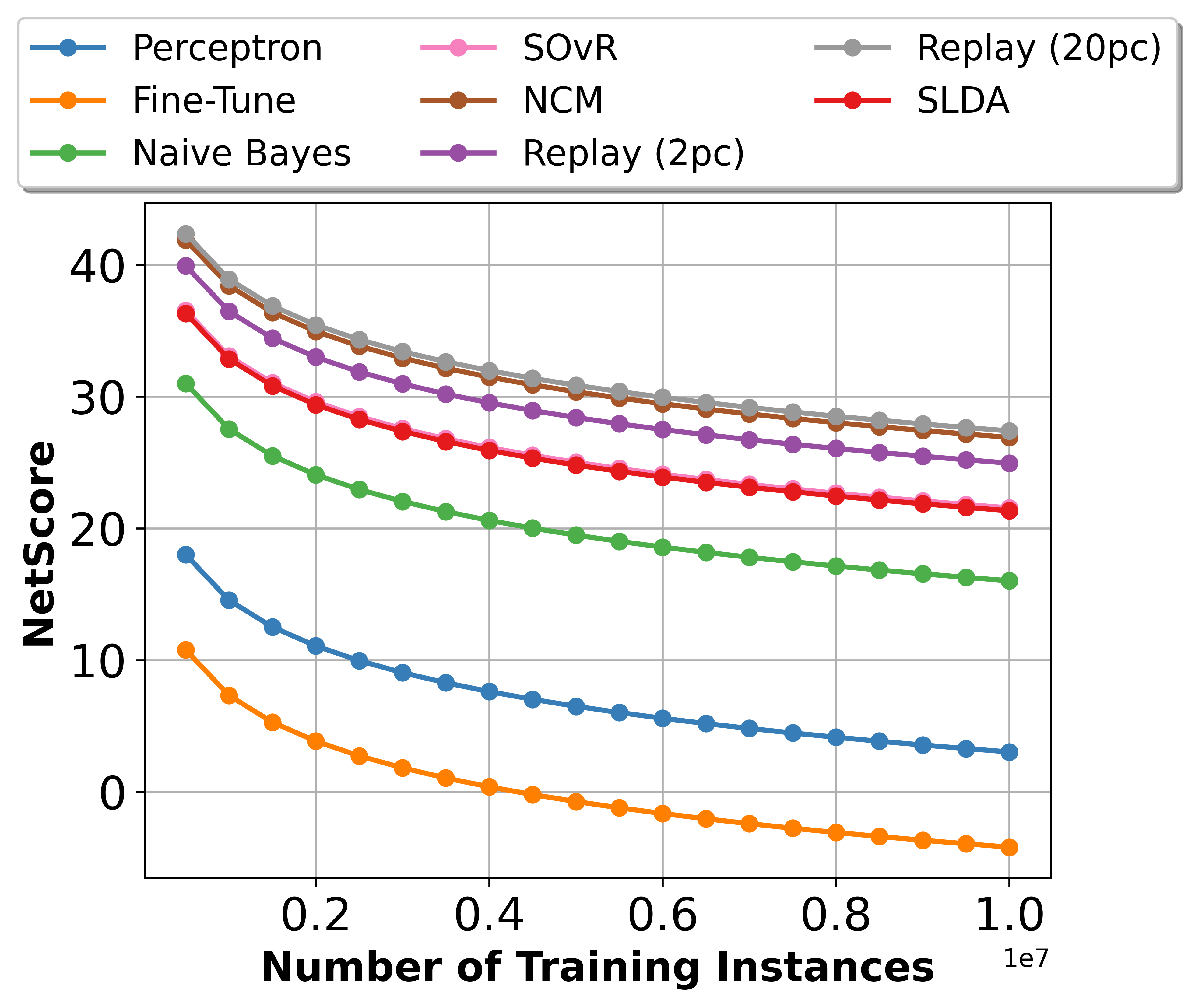}
        \caption{EfficientNet-B1}
    \end{subfigure} %
    \centering
    \begin{subfigure}[t]{0.32\linewidth}
        \includegraphics[width=\linewidth]{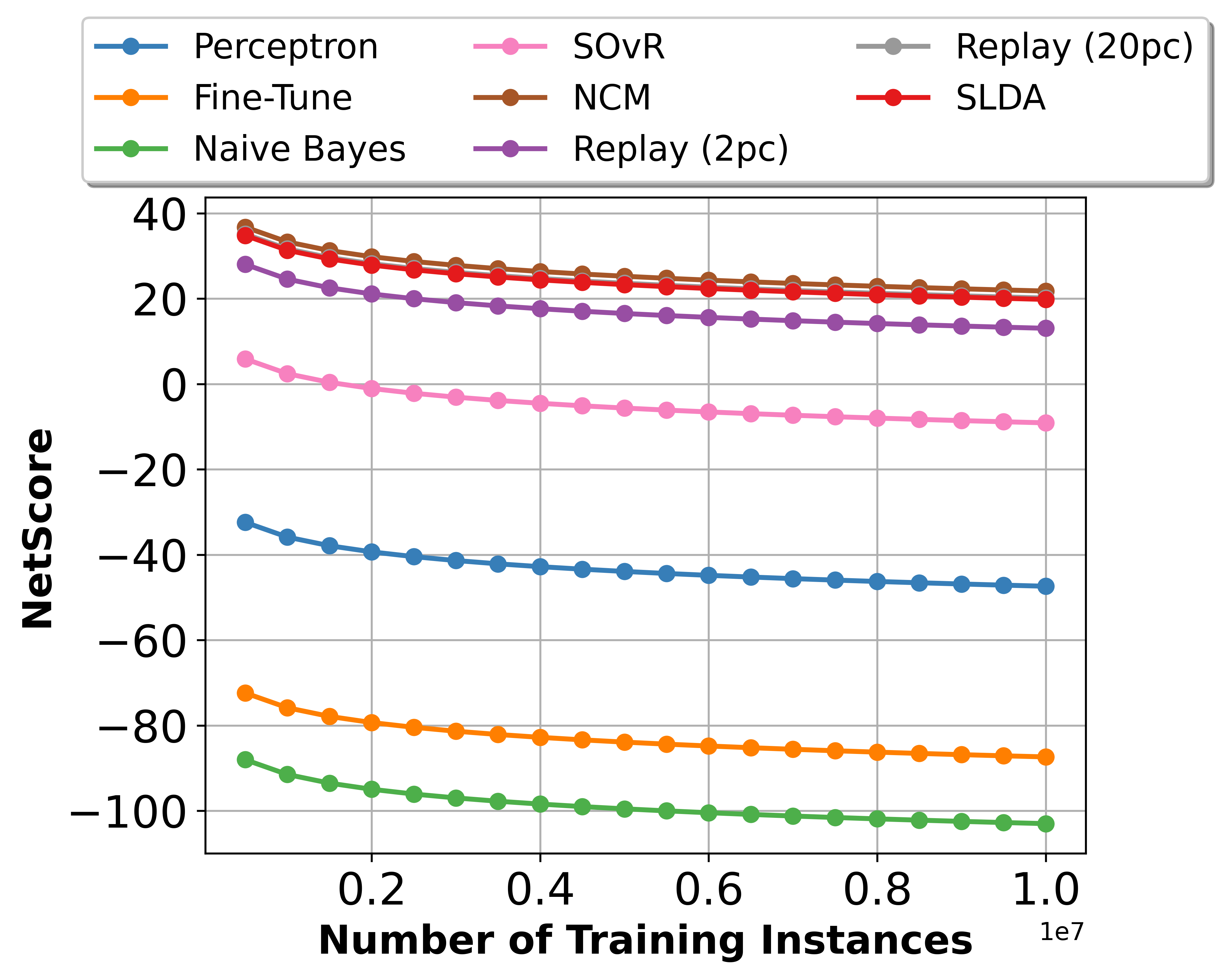}
        \caption{ResNet-18}
    \end{subfigure} %
    \caption{NetScore values interpolated to account for additional samples in a dataset for each backbone CNN and continual learner.
    }
    \label{fig:netscore-scale-instance-plots}
\end{figure*}

\begin{figure*}[t]
    \centering
    \begin{subfigure}[t]{0.32\linewidth}
        \includegraphics[width=\linewidth]{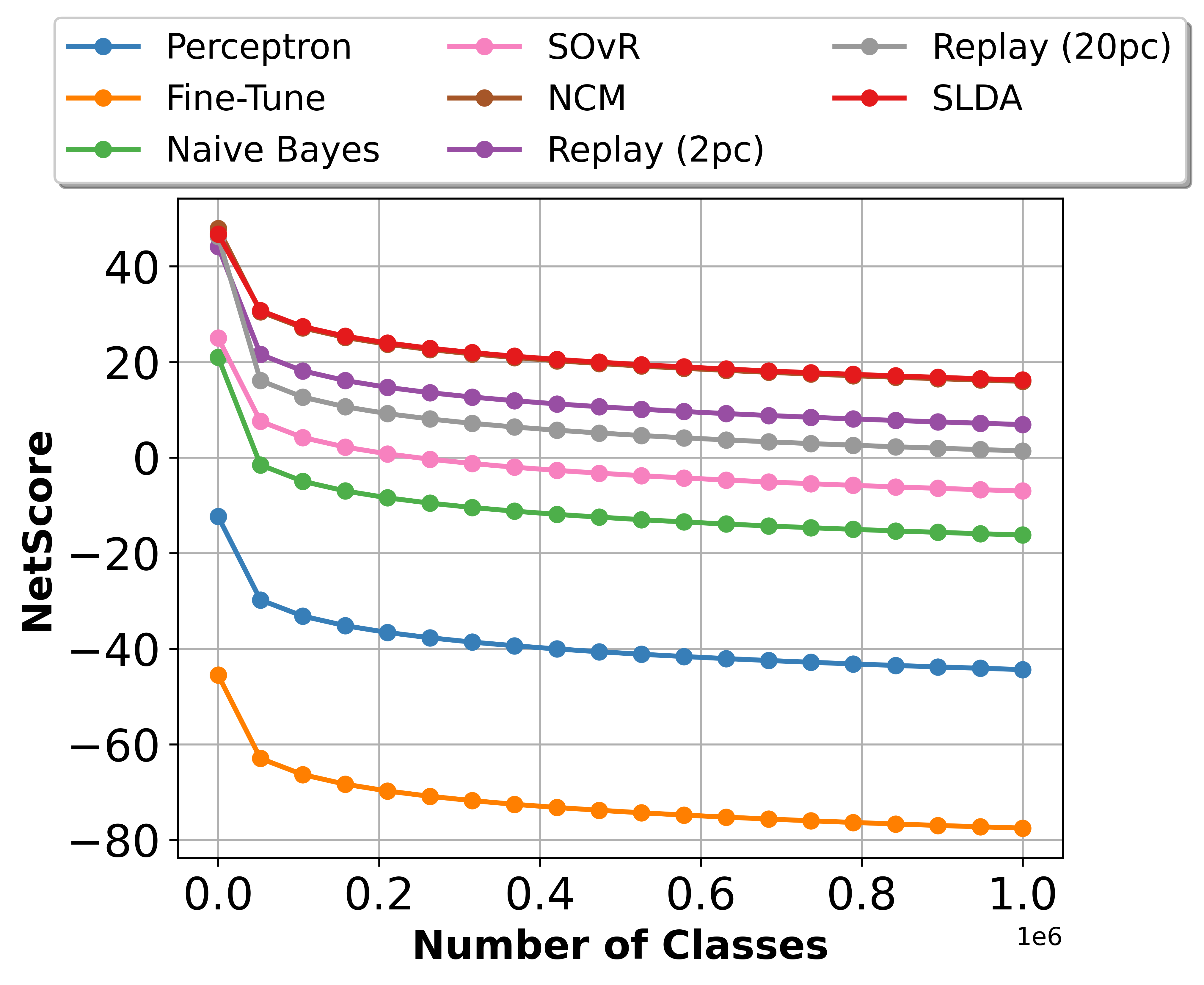}
        \caption{MobileNet-v3 (Small)}
    \end{subfigure} %
    \centering
    \begin{subfigure}[t]{0.32\linewidth}
        \includegraphics[width=\linewidth]{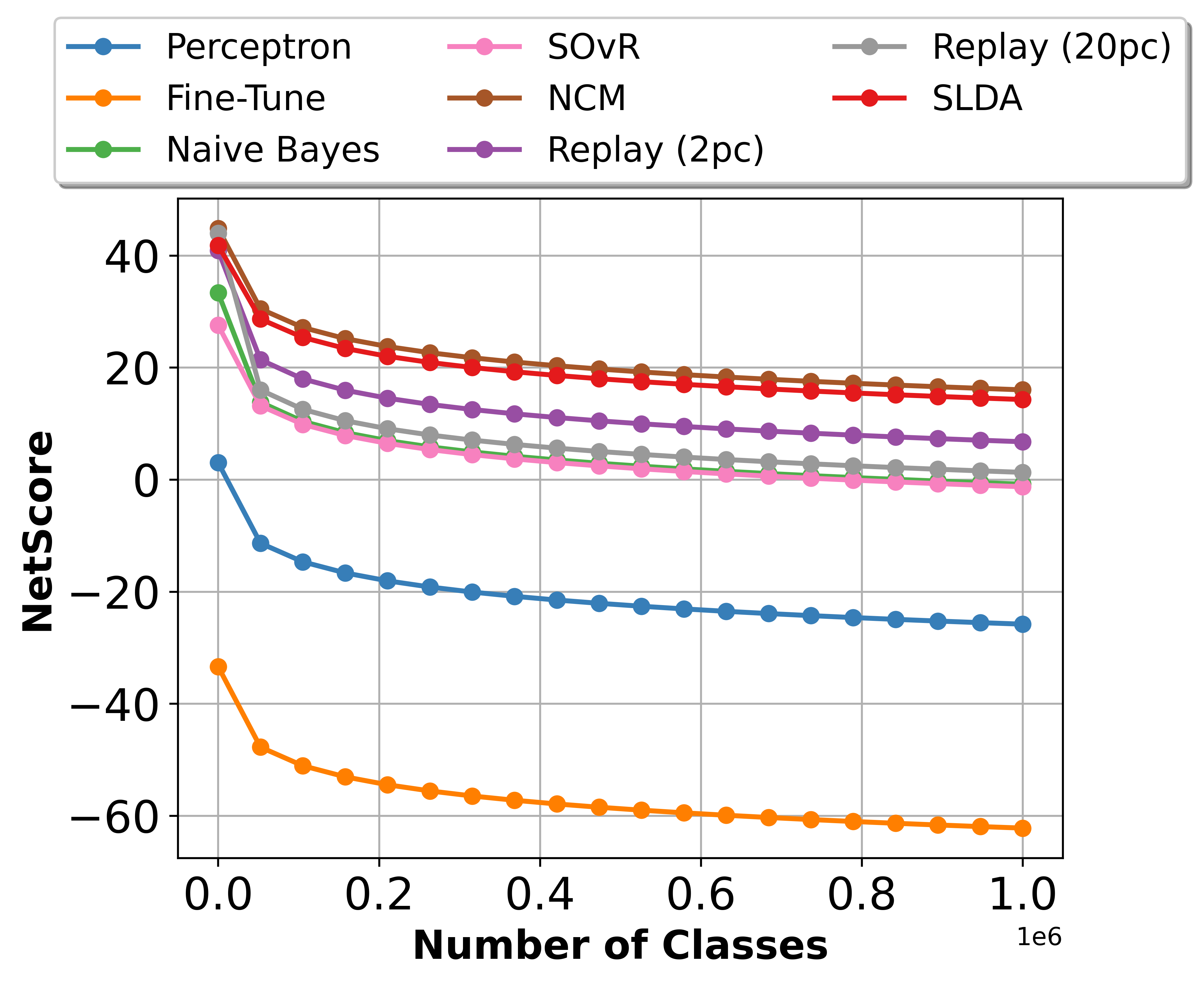}
        \caption{MobileNet-v3 (Large)}
    \end{subfigure} %
    \centering
    \begin{subfigure}[t]{0.32\linewidth}
        \includegraphics[width=\linewidth]{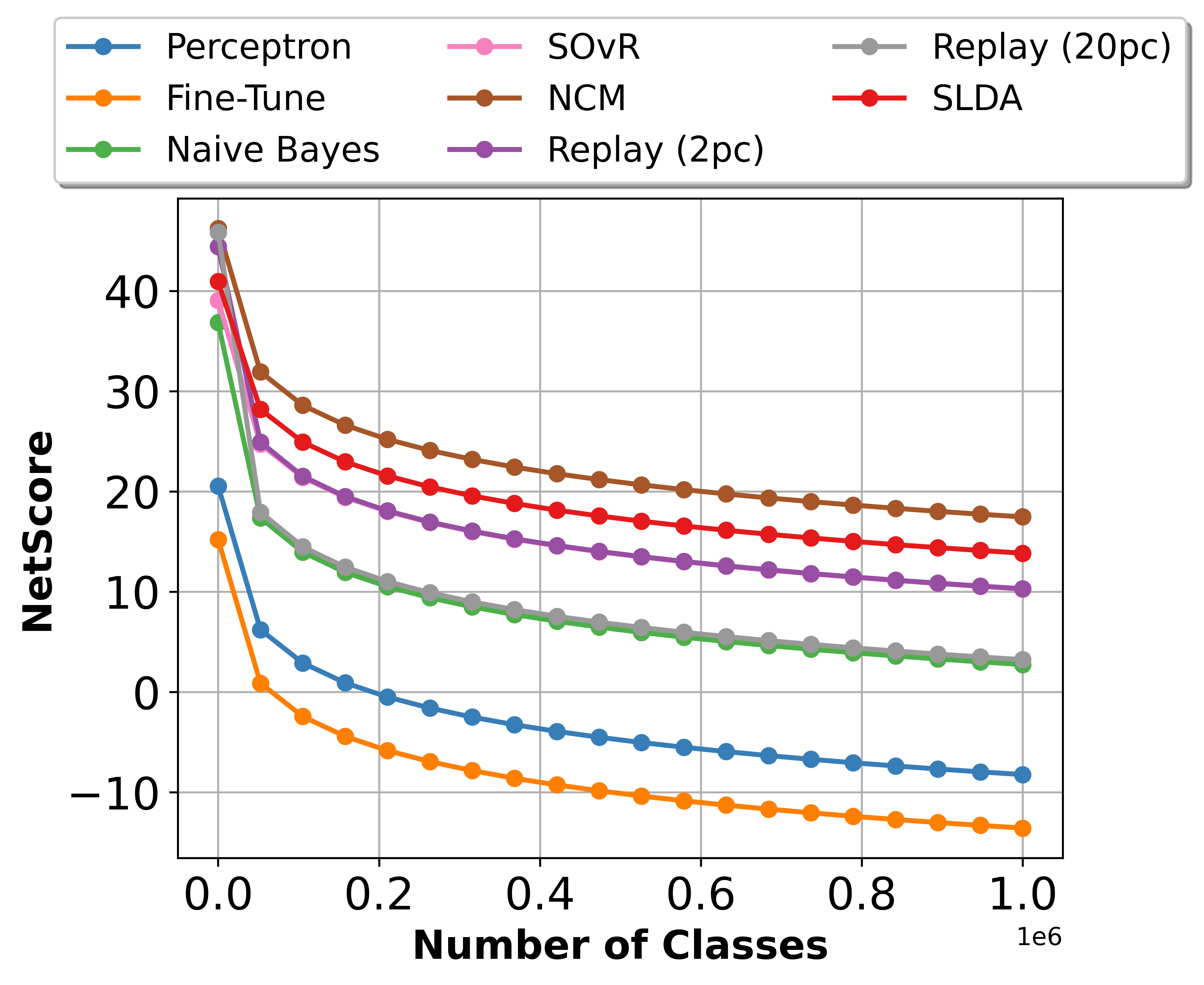}
        \caption{EfficientNet-B0}
    \end{subfigure} %
    \\
    \centering
    \begin{subfigure}[t]{0.32\linewidth}
        \includegraphics[width=\linewidth]{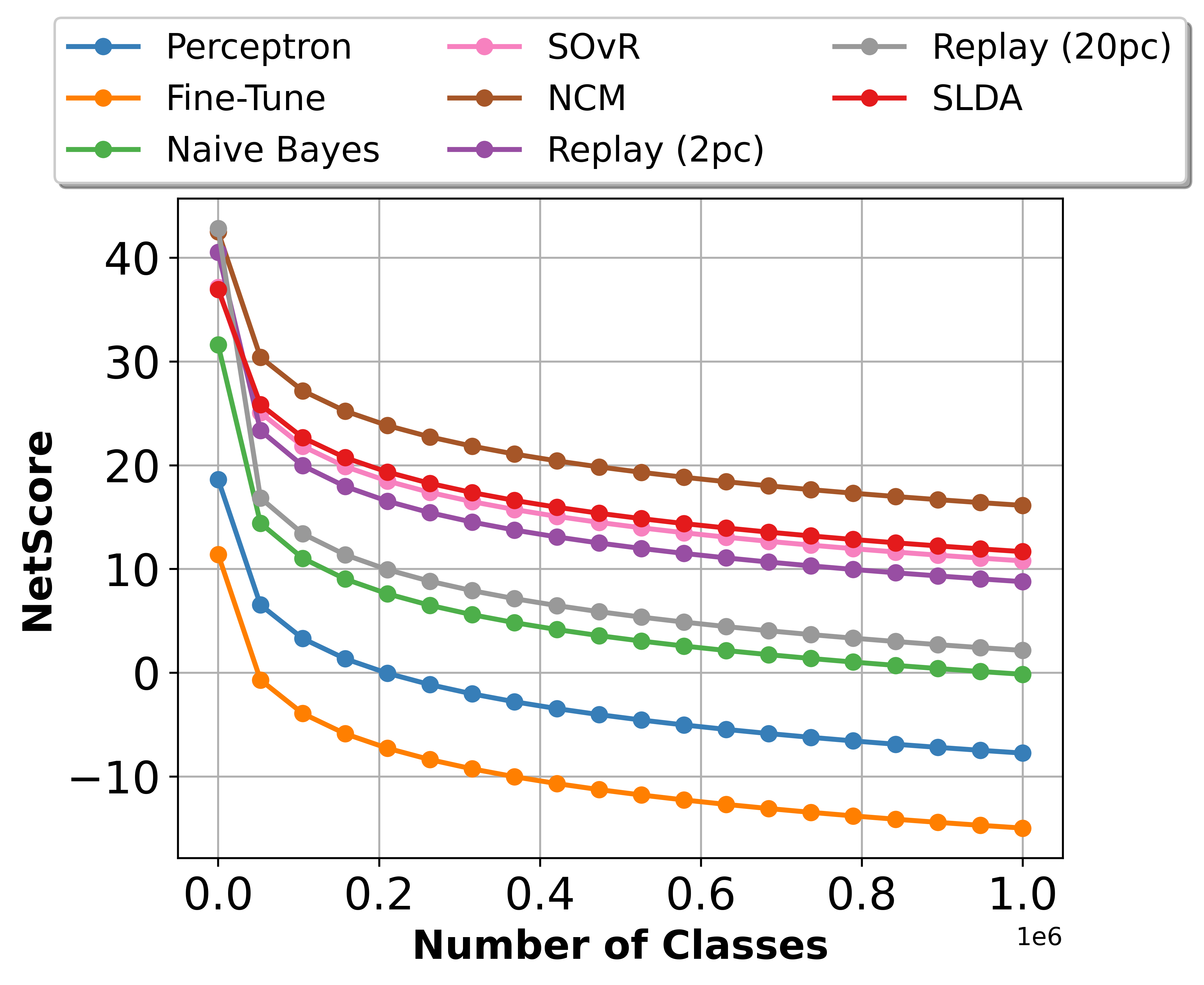}
        \caption{EfficientNet-B1}
    \end{subfigure} %
    \centering
    \begin{subfigure}[t]{0.32\linewidth}
        \includegraphics[width=\linewidth]{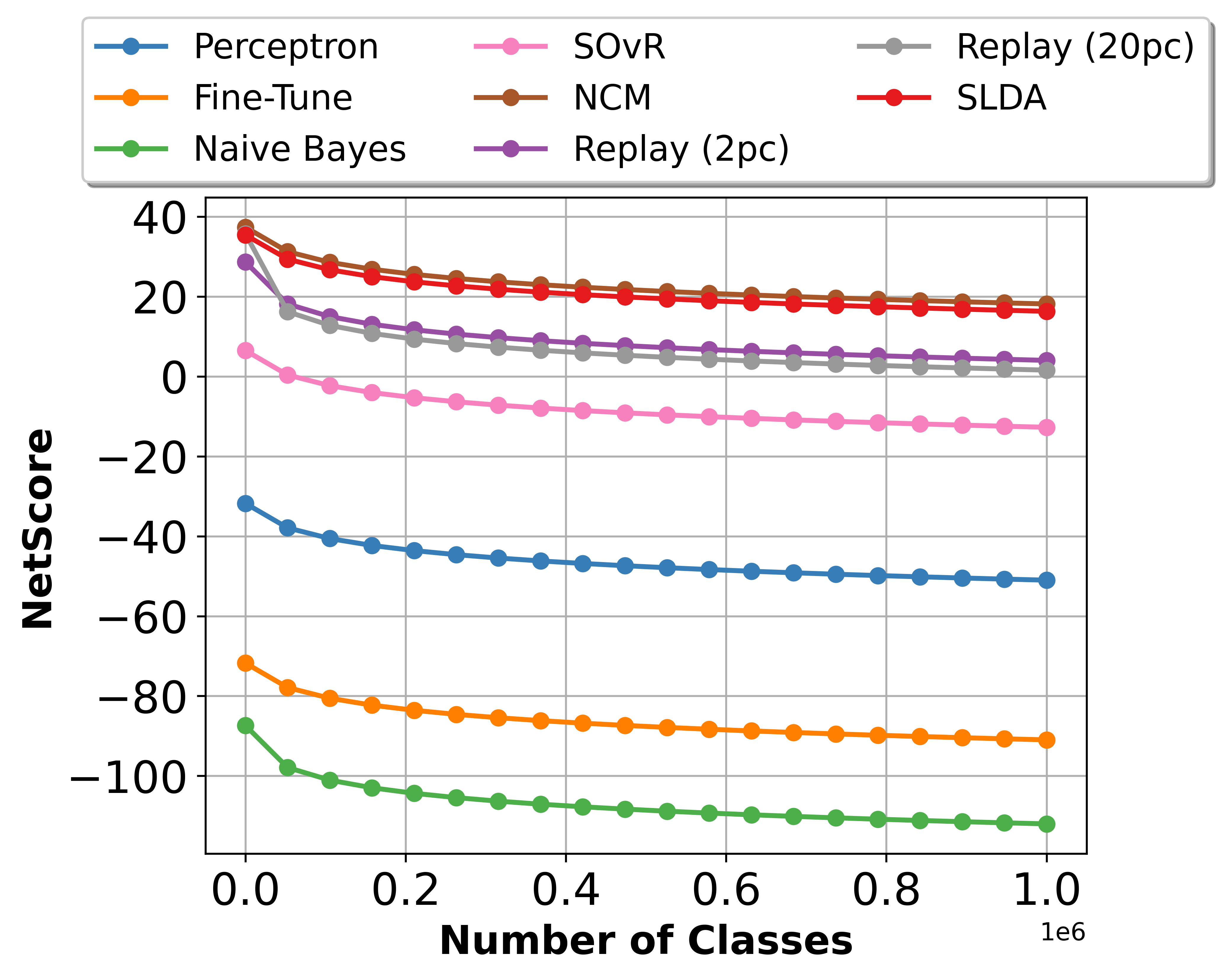}
        \caption{ResNet-18}
    \end{subfigure} %
    \caption{NetScore values interpolated to account for additional classes in a dataset for each backbone CNN and continual learner.
    }
    \label{fig:netscore-scale-class-plots}
\end{figure*}

Next, we examine how robust the NetScore metric is to changes in dataset scale. We provide high-level proof-of-concept plots to examine how the NetScore metric scales when a dataset contains more samples and when a dataset contains more classes by interpolating our NetScore results from Table~\ref{tab:open-loris-summarized-results}. Specifically, to understand how NetScore scales with the number of samples in a dataset, we use the original classification efficacy and memory (parameters) values from Table~\ref{tab:open-loris-summarized-results}, but scale the compute (run-time) by multiples greater than one, e.g., scaling compute time by a factor of two would give us the compute time required to run the experiment on a dataset two times the size of the original dataset containing 440,919 train and 53,295 test samples. Similarly, to understand how NetScore scales with the number of classes, we use the original classification efficacy and compute (run-time) values from Table~\ref{tab:open-loris-summarized-results}, but compute the number of parameters (memory) required by each model for datasets containing more classes. The resulting plots are in Fig.~\ref{fig:netscore-scale-instance-plots} and in Fig.~\ref{fig:netscore-scale-class-plots} for datasets with more samples and more classes, respectively. As expected, when the number of samples or number of classes in a dataset increases, the online continual learner NetScore values monotonically decrease. However, some methods are more negatively impacted by the addition of more classes (e.g., replay) than others.








\subsubsection{F-SIOL-310}

\begin{table}[h]
\caption{Final accuracy (\%) results on the F-SIOL-310 dataset with the class-iid data ordering under the 5-shot and 10-shot learning scenarios. Each result is the average over 3 runs with different permutations of the data. We format the \underline{\textbf{first}}, \textbf{second} and \underline{third} best overall results. \label{tab:fsiol-results}}
\centering
\resizebox{\textwidth}{!}{%
\begin{tabular}{lcccccccccccc}
\toprule
& \multicolumn{6}{c}{\textsc{5-Shot}} & \multicolumn{6}{c}{\textsc{10-Shot}} \\
\cmidrule(r){2-7} \cmidrule(r){8-13}
\textsc{Method} & \textsc{MNet-S} & \textsc{MNet-L} & \textsc{ENet-B0} & \textsc{ENet-B1} & \textsc{RN-18} & \textsc{Mean} & \textsc{MNet-S} & \textsc{MNet-L} & \textsc{ENet-B0} & \textsc{ENet-B1} & \textsc{RN-18} & \textsc{Mean} \\ 
\midrule
Perceptron & 19.4 & 21.6 & 35.0 & 39.5 & 5.0 & 24.1 & 12.7 & 18.7 & 34.4 & 51.0 & 5.3 & 24.4 \\
Fine-Tune & 18.0 & 22.9 & 41.5 & 48.4 & 9.0 & 28.0 & 12.4 & 19.8 & 36.2 & 43.8 & 8.2 & 24.1 \\
Naive Bayes & 26.7 & 50.7 & 79.2 & 82.9 & 3.9 & 48.7 & 34.8 & 56.4 & 79.2 & 85.2 & 1.8 & 51.5 \\
SOvR & 54.8 & 66.0 & 65.9 & 71.5 & 44.8 & 60.6 & 56.2 & 68.2 & 65.0 & 80.6 & 49.7 & 63.9 \\
CBCL & 82.9 & 85.8 & 83.9 & 80.2 & 81.5 & \underline{82.9} & 90.9 & 91.8 & 91.5 & 90.4 & 89.9 & \underline{90.9} \\
NCM & 82.9 & 85.9 & 85.3 & 86.2 & 85.0 & \textbf{85.1} & 90.9 & 91.8 & 91.9 & 92.0 & 90.8 & \textbf{91.5} \\
Replay (2pc) & 52.0 & 58.5 & 56.7 & 59.9 & 59.2 & 57.3 & 63.3 & 73.0 & 74.9 & 76.1 & 75.8 & 72.6 \\
SLDA & 88.5 & 90.7 & 90.3 & 89.0 & 85.4 & \underline{\textbf{88.8}} & 93.9 & 96.6 & 96.6 & 94.7 & 91.5 & \underline{\textbf{94.7}} \\
\bottomrule
\end{tabular}
}
\end{table}

Finally, we were interested to understand how each of the continual learning methods performed on a dataset designed specifically for few-shot continual learning. Specifically, we use the F-SIOL-310 (Few-Shot Incremental Object Learning) dataset~\citep{ayub2021fsiol}, which consists of 620 total static images of 310 unique object instances from 22 unique classes. To perform online continual learning experiments using this dataset, we use the class-iid data ordering in both 5-Shot and 10-Shot learning scenarios, as suggested by \citet{ayub2021fsiol}. For the 5-Shot learning scenario, we randomly select five images from each class for training and the rest of the dataset is used for testing. For the 10-Shot learning scenario, we randomly select 10 images from each class for training and the rest of the dataset is used for testing. We run each experiment three times with different permutations of the classes and report the average results over the three runs. We use the parameter settings from Sec.~\ref{sec:implementation} for all methods except replay. For replay, we use a maximum buffer size of 44 as suggested by \citet{ayub2021fsiol}.

In addition to the online continual learners studied in the rest of this paper, we also study an online variant of the \textbf{Centroid-Based Concept Learning} (CBCL) model designed specifically for few-shot continual learning~\citep{ayub2020cognitively}. CBCL updates multiple centroids per class during training. Specifically, the first time a sample from a class is seen, it is stored as a centroid. After that, each time the model sees an example from a class, it computes the distance to the nearest centroid from that class. If the sample is within a threshold distance of the nearest centroid, it is merged with that centroid, otherwise it forms a new centroid. To perform inference, CBCL performs a weighted nearest neighbor computation where the class weight is assigned as the inverse of the number of examples that have been seen for a particular class. To make the algorithm amenable to online learning, we perform centroid updates on a sample-by-sample basis using Eq.~\ref{eq:running-mean}, we assign class weights during testing according to the number of samples that have been seen for a class, and we merge the two closest class clusters into one cluster when the number of centroids needs to be reduced as:
\begin{equation}
    \mathbf{w}_{i} = \frac{\mathbf{w}_{i} +\mathbf{w}_{j}}{c_{i}+c_{j}} \enspace, \qquad  c_{i} = c_{i} + c_{j} \enspace ,
\end{equation}
where $\mathbf{w}_{i}$ and $\mathbf{w}_{j}$ are the two closest clusters with associated counts $c_{i}$ and $c_{j}$. We use the parameters suggested by \citet{ayub2021fsiol}, i.e., a distance threshold of 17, a nearest neighbor $k$ value of 1, and a maximum buffer size of 44 centroids.

The results for each continual learner with each backbone CNN are in Table~\ref{tab:fsiol-results}. Overall, we found that the top-performing method across both the 5-Shot and 10-Shot learning scenarios was SLDA. The next top performers were NCM and CBCL. All three of these methods are distance-based classifiers, which have been shown to work well in low-shot learning settings~\citep{snell2017prototypical,vinyals2016matching}. Although CBCL performed fairly well, it requires more memory and compute than NCM to update multiple centroids and has slightly worse performance, so NCM is still a strong method for this dataset. While the replay method was a top performer on other datasets, it struggles in the few-shot learning setting, which is likely due to not having enough replays of previous examples to mitigate forgetting. Based on these results, we recommend using distance-based classifiers for low-shot embedded continual learning applications. Specifically, it could be interesting to consider combinations of these methods in future work such as the combination of SLDA with multiple centroids as in CBCL.

\subsubsection{Overall Results}

\begin{figure}[t]
\begin{center}
    \includegraphics[width=0.6\textwidth]{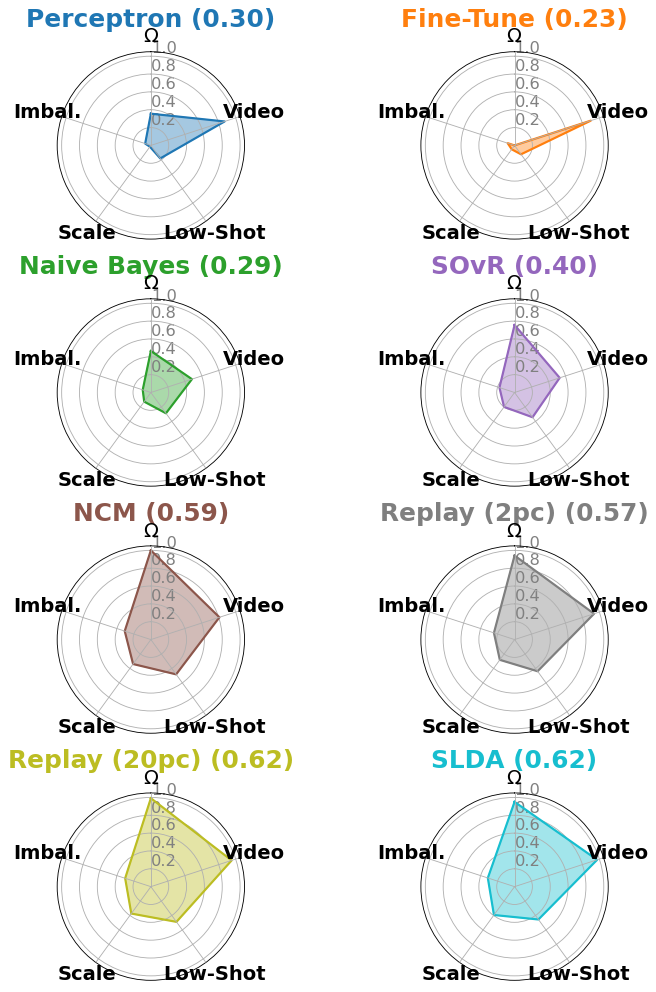}
\end{center}
\caption{Spider plots comparing the normalized performance of each online continual learner with respect to NetScore ($\Omega$), the ability to learn from temporally correlated videos (Video), the ability to generalize from few labeled inputs (Low-Shot), the ability to scale to large-scale data (Scale), and the ability to perform well on imbalanced data (Imbal). The average performance of each learner across all of these axes is at the top of each plot. Each metric has been averaged across all CNNs and normalized to fall in the range $[0,1]$.}
\label{fig:spider}
\end{figure}

To better understand the performance of individual methods across experiments, we create spider plots with the performance of each online continual learner with respect to NetScore, the ability to learn from temporally correlated videos, the ability to generalize from few labeled inputs, the ability to scale to large-scale data, and the ability to perform well on imbalanced data. To create these plots, we first average the performance of each continual learner across all five backbone CNNs. We then normalize NetScore values to fall in the range $[0,1]$ such that the method with the best NetScore has a normalized score of 1 and the method with the worst NetScore has a normalized score of 0. For learner performance on videos, low-shot learning, ability to scale, and ability to learn from imbalanced data, we report the final scores achieved on the OpenLORIS (instance), OpenLORIS (low-shot instance), Places-365 (harmonic mean), and Places-LT (harmonic mean) datasets, respectively. The resulting spider plots are in Fig.~\ref{fig:spider}, where the title of each plot contains the name of the online continual learner and its associated average performance across the five axes considered in the plots.

Overall, these plots indicate that SLDA and replay (20pc) performed the best across all five axes. NCM and replay (2pc) also had strong overall performance. While the fine-tune and perceptron methods performed the worst overall, they had strong performance on the video-based OpenLORIS dataset. In contrast, SOvR had better performance overall, but worse performance compared to fine-tune and perceptron on videos.

\subsection{Online Continual Learner Considerations}
\label{sec:considerations}

There are several strengths and weaknesses of each online continual learning method that must be considered when choosing methods for embedded applications. We highlight the strengths and weaknesses of each online continual learning method next.

\textbf{Fine-Tune} and \textbf{Perceptron} - Strengths: fast; only require memory to store the classifier; work well on iid data. Weaknesses: do not have mechanisms to mitigate catastrophic forgetting (i.e., do not work well across all data orderings); do not perform well on imbalanced data since they will likely overfit to overrepresented classes and underfit to underrepresented classes; fine-tune has many hyperparameters to tune (e.g., learning rate, weight decay, momentum, etc.).

\textbf{SOvR} - Strengths: running means are order agnostic; no hyperparameters to tune. Weaknesses: does not work well in many of the experimental settings evaluated in this paper.

\textbf{Naive Bayes} - Strengths: assumes different variances per class so it can model individual class distributions; fairly robust to data orderings since its running mean vectors are order agnostic and its class variance vectors will result in (at most) gradual forgetting; it is a generative classifier which could allow it to better model data from very few examples (i.e., low-shot settings). Weaknesses: assumes features are independent, which isn’t always guaranteed; assumes data is Gaussian, which isn’t always guaranteed.

\textbf{NCM} - Strengths: class mean vectors are order agnostic; no hyperparameters to tune; simple to implement. Weaknesses: cannot model outlier data.

\textbf{SLDA} - Strengths: fairly robust to data orderings since its running mean vectors are order agnostic and its covariance matrix will result in (at most) gradual forgetting; it is a generative classifier which could allow it to better model data from very few examples (i.e., low-shot settings); works well across a variety of experiments. Weaknesses: assumes data is Gaussian, which isn’t always guaranteed; assumes classes have equal covariances, which isn’t always guaranteed.

\textbf{Replay} - Strengths: fairly robust to data orderings since previous data is maintained in a memory buffer; works well across a variety of experiments. Weaknesses: requires storage of additional memory to work well; has many hyperparameters to tune (e.g., buffer size, replay selection method, learning rate, weight decay, momentum, etc.).

\end{document}